\pdfoutput=1

\documentclass[11pt]{article}

\usepackage[final]{acl}

\usepackage{times}
\usepackage{latexsym}

\usepackage[T1]{fontenc}

\usepackage[utf8]{inputenc}

\usepackage{microtype}

\usepackage{inconsolata}

\usepackage{graphicx}

\usepackage{arydshln}

\usepackage{algorithm} 
\usepackage{algpseudocode}

\usepackage[skins]{tcolorbox}
\usepackage{graphicx}
\usepackage{enumitem}
\usepackage{makecell}

\usepackage{booktabs}
\usepackage{multirow}

\newcolumntype{C}[1]{>{\centering\arraybackslash}m{#1}}

\tcbuselibrary{listingsutf8}
\global

\tcbset{
  aibox/.style={
    width=474.18663pt,
    top=10pt,
    colback=white,
    colframe=black,
    colbacktitle=black,
    enhanced,
    center,
    attach boxed title to top left={yshift=-0.1in,xshift=0.15in},
    boxed title style={boxrule=0pt,colframe=white,},
  }
}
\newtcolorbox{AIbox}[2][]{aibox,title=#2,#1}

\usepackage{amsmath}

\usepackage{graphicx}
\usepackage{cleveref}
\usepackage{booktabs}
\usepackage{subcaption}
\usepackage{array}

\usepackage{soul}

\usepackage{caption}
\usepackage{subcaption}

\usepackage{booktabs}
\usepackage{multirow}
\usepackage{makecell}

\usepackage{enumitem}
\usepackage{amsmath}
\usepackage{svg}

\usepackage{tabularx}             

\usepackage{annotation}
\usepackage{tikz}
\usepackage{hyperref}

\definecolor{pastelBlue}{RGB}{174,198,207}   
\definecolor{pastelRed}{RGB}{255,179,186}      
\definecolor{pastelOrange}{RGB}{255,223,186}   
\definecolor{pastelGreen}{RGB}{203,240,191}    
\definecolor{pastelViolet}{RGB}{213,181,226}   
\definecolor{pastelYellow}{RGB}{255,255,204}   
\definecolor{pastelPurple}{RGB}{230,230,250}   
\definecolor{pastelPink}{RGB}{255,192,203}     

\usepackage{pgfplots}
\usetikzlibrary{fillbetween} 
\pgfplotsset{compat=1.17}

\newcommand{\dataset}{FAME}
\newcommand{\pipeline}{MIMIC}

\title{You need to MIMIC to get FAME: \\Solving Meeting Transcript Scarcity with Multi-Agent Conversations}

\author{Frederic Kirstein\textsuperscript{1,2}, Muneeb Khan\textsuperscript{1}, Jan Philip Wahle\textsuperscript{1}, Terry Ruas\textsuperscript{1}, Bela Gipp\textsuperscript{1} \\
  \textsuperscript{1}University of Göttingen, Germany \\
\textsuperscript{2}\texttt{kirstein@gipplab.org} }

\begin{document}
\maketitle
\AddAnnotationRef

\begin{abstract}
\label{sec:data_insights}

Meeting summarization suffers from limited high-quality data, mainly due to privacy restrictions and expensive collection processes.
We address this gap with \dataset{}, a dataset of 500 meetings in English and 300 in German produced by \pipeline{}, our new multi-agent meeting synthesis framework that generates meeting transcripts on a given knowledge source by defining psychologically grounded participant profiles, outlining the conversation, and orchestrating a large language model (LLM) debate.
A modular post-processing step refines these outputs, mitigating potential repetitiveness and overly formal tones, ensuring coherent, credible dialogues at scale.
We also propose a psychologically grounded evaluation framework assessing naturalness, social behavior authenticity, and transcript difficulties.
Human assessments show that \dataset{} approximates real-meeting spontaneity (4.5/5 in naturalness), preserves speaker-centric challenges (3/5 in spoken language), and introduces richer information-oriented difficulty (4/5 in difficulty).
These findings highlight that \dataset{} is a good and scalable proxy for real-world meeting conditions.
It enables new test scenarios for meeting summarization research and other conversation-centric applications in tasks requiring conversation data or simulating social scenarios under behavioral constraints\footnote{Resources are available as per \Cref{sec:Data_Availability} on \href{https://github.com/FKIRSTE/synthetic_meeting_transcript}{GitHub}.}.
\end{abstract}

\begin{figure*}
    \centering
    \includegraphics[width=\textwidth]{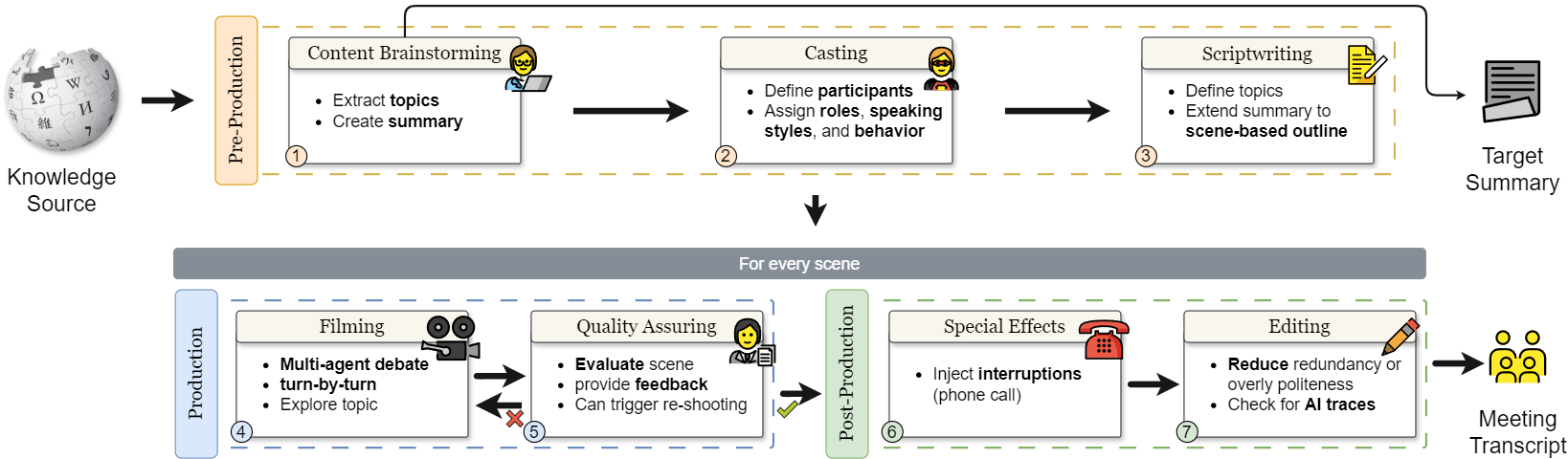}
    \caption{Architecture of our meeting generation framework \pipeline{}. Stages 1 to 3 are realized as a linear pipeline of LLM instances. Stage 4 is an agent orchestrating participants played by agents, who can decide on the length of the meeting through voting. Stage 5 checks and optionally provides feedback for refinement to stage 4 agent. The remaining stages, 6 and 7, polish the generated conversation.}
    \label{fig:main_figure}
\end{figure*}

\section{Introduction}
Meetings underlie collaboration and decision-making in corporations, academia, and government.
Meeting summaries help record key discussion points, update absentees, and capture to-dos \cite{ZhongYYZ21h, HuGDD23a, LaskarFCB23}.
While AI-based summaries are available on platforms such as Zoom\footnote{\href{https://www.zoom.com/en/ai-assistant}{https://www.zoom.com/en/ai-assistant}}, Microsoft Teams\footnote{\href{https://copilot.cloud.microsoft}{https://copilot.cloud.microsoft}}, they typically build on limited English data that does not represent the diversity of real meetings, e.g., multilingual sessions, specialized discussions \cite{KirsteinWGR25}.
Data scarcity for training and testing meeting summarization systems persists due to privacy and intellectual property concerns, along with expensive manual annotation \cite{AbachaYFL23}.
Existing corpora, such as AMI \cite{MccowanCKA05}, ICSI \cite{JaninBEE03}, and MeetingBank \cite{HuGDD23a}, offer only a narrow range of scenarios, which primarily revolve around staged business, academic, or parliamentary meetings.
Non-English resources like FREDSum \cite{RennardSGH23a}, containing manually transcribed and annotated political debates, remain sporadic and underscore the lack of linguistic diversity.

Researchers have explored synthetic transcripts to address this scarcity, but many methods are sub-optimal in scalability and realism.
Single-model, omniscient continuation \cite{QiuP24, ZhouSEK24} can produce dialogues lacking actual knowledge and speaker interplay, while crowdsourced role-plays are expensive \cite{MccowanCKA05,ThulkeGJD24}. 
Automated heuristics (e.g., noising, swapping) often yield disjointed conversations \cite{ChenY21, ParkSL22, LiuMSN22}. 
These approaches struggle to balance large-scale generation with authentic group dynamics and credible topic evolution, even though they can process and generate thousands of tokens.

We introduce \textbf{\pipeline{}} (\textbf{M}ulti-agent \textbf{IMI}tation of \textbf{C}onversations, see \Cref{fig:main_figure}), a movie-production-inspired framework based on multi-agent debate \cite{LiangHJW24a,DuLTT24}.
\pipeline{} summarizes a knowledge source, expands the summary into an agenda, and orchestrates psychologically grounded agents debating turn-by-turn, allowing interruptions (e.g., phone calls).
A modular refinement step mitigates repetitions or overly formal speech, ensuring plausible and coherent discourse at scale.
With \pipeline{}, we generate \textbf{\dataset{}} (\textbf{FA}ke \textbf{ME}etings), a corpus of 800 meetings (500 English, 300 German parallel in input) covering 14 meeting types (e.g., project updates, brainstorming) and 300 Wikipedia articles as a knowledge source.

We evaluate the synthetic meetings on quality criteria, e.g., naturalness, coherence, \cite{ChenPTK23}, and transcript difficulties, e.g., scattered information \cite{KirsteinWRG24a}, and propose a new psychology-based measure of behavioral authenticity \cite{ChoiAVQ20}.
Human ratings confirm that \dataset{} achieves near-real spontaneity (4.5/5  in naturalness), preserves speaker-related challenges (3/5 in spoken language), and intensifies low-information-density difficulties (4/5).
Comparisons with real transcripts and 100 crowdsourced meeting experiences show a similar behavioral pattern.
Evaluations of GPT-4o \cite{OpenAIAAA24}, Gemini 1.5 pro \cite{GeminiTeamRST24}, DeepSeek-R1 Distill Llama 70B \cite{DeepSeekAIGYZ25a}, Llama 3.3 70B\footnote{We will refer to these as GPT, Gemini, DeepSeek, and Llama throughout the paper.} \cite{GrattafioriDJP24} reveal persistent context-handling issues. 
Ablation studies show that \pipeline{} reliably generates good transcripts using different inputs or LLMs.

\paragraph{Contributions.}
\begin{itemize}[noitemsep, topsep=0pt, leftmargin=*]
    \item \textbf{\pipeline{}}: A multi-agent simulation method that captures realistic group dynamics.
    \item \textbf{\dataset{}}: A corpus of 800 meetings in English (500) and German (300) on diverse topics and meeting formats, including quality annotations.
    \item A \textbf{psychology-based evaluation framework} for behavioral authenticity, addressing a gap in measuring agent interactions. 
\end{itemize}

\section{Related Works}
\noindent
\paragraph{Meeting summarization datasets.}
Most meeting summarization research relies on a few standard English-only corpora \cite{KirsteinWGR25}, notably AMI \cite{MccowanCKA05}, i.e., staged business, and ICSI \cite{JaninBEE03}, i.e., academic.
While QMSum \cite{ZhongYYZ21h} and MeetingBank \cite{HuGDD23a} broaden coverage to parliamentary sessions and city councils, non-English data remains sparse (e.g., FREDSum \cite{RennardSGH23a} for French and ELITR \cite{NedoluzhkoSHG22} for Czech).
We address this gaps by releasing \dataset{}, a corpus of 800 synthetic English and German meetings spanning diverse topics, meeting formats, and speaking styles.

\paragraph{Synthetic dialogue generation.}
Existing synthetic meeting data generation approaches often rely on relatively simple text continuation by LLMs \cite{QiuP24} or heuristics, e.g., noising, swapping \cite{ChenY21, ParkSL22, LiuMSN22}, risking superficial turn-taking and unrealistic participant behavior \cite{KirsteinWGR25}.
Small-scale manual simulations \cite{ThulkeGJD24} offer greater realism but are costly and hard to scale.
Recent tools like Google’s NotebookLM \cite{google-2024} and Nvidia’s PDF-to-Podcast \cite{nvidia-2025} transform documents into two-speaker podcasts but lack multi-participant group dynamics.
In contrast, \pipeline{} simulates turn-by-turn interactions among psychology-based agents with their own memory, allowing spontaneous debates and evolving stances. 
A post-processing module addresses common LLM flaws (e.g., repetition, vocabulary), ensuring high-quality, naturally flowing conversations.


\section{The \pipeline{} Methodology}
\label{sec:pipeline}

We propose \pipeline{}, a multi-agent framework inspired by movie production to generate synthetic meeting transcripts from a knowledge source through multi-agent debate (see \Cref{fig:main_figure}).
The basic idea is to summarize the knowledge source to distill its key highlights to be discussed in a multi-agent-LLM setup.
This setup emulates a discussion among participants with distinct personas and their private memory of the knowledge source, including real-world dynamics such as turn-taking, disagreements, clarifications, and topic continuity.
\pipeline{} operates in three phases, i.e., \textit{pre-production}, \textit{production}, and \textit{post-production}, with seven stages overall (splitted 3/2/2 between each phase).
We explain here the methodological background of \pipeline{}, the corresponding prompts, and implementation details are covered in \Cref{sec:appendix_A,sec:appendix_C}.
In \Cref{sec:appendix_B}, we present the result of each stage using an example knowledge source.

\subsection{Pre-Production Phase}
This phase establishes foundational elements, including the meeting’s target summary, participant roles, and an agenda-like outline.

\paragraph{Stage 1: Content Brainstorming.}
Given a source text, we prompt an LLM to extract hierarchical topics and subtopics, following the approach outlined in \citet{Paoli23a}.
An LLM composes an abstractive target summary following \citet{GaoJYZ24}, guided by five human-written QMSum summaries for consistent brevity, style, and structure.
This summary covers the later discussion points and topic flow, acting as a basic outline.

\paragraph{Stage 2: Casting.}
\label{sec:stage_2_casting}

We define participant profiles suited to the meeting context.
Each profile contains a functional role (e.g., project manager, technical expert), background (e.g., experience, qualifications), domain expertise, and a distinct perspective (e.g., favoring practical solutions).
An LLM iteratively creates these profiles, ensuring complementary viewpoints without redundancy.
Next, each profile receives a speaking style, including tone (e.g., formal), language complexity (e.g., jargon), communication style (e.g., assertive), plus filler words (e.g., ``um,'' ``you know'') and catchphrases to align the person's language with their role.

We distributed select knowledge-source paragraphs to each participant based on expertise \cite{LiWXG25}, introducing knowledge imbalances that foster reliance on one another.
Building on \citet{SerapioGarciaSCS23a} research on LLM personality traits, each participant is assigned psychologically grounded behaviors, e.g., evaluator-critic, blocker \cite{BenneS48}, drawn from a curated list (full list given in \Cref{tab:app_role_overview} in \Cref{sec:appendix_C}).
These behaviors can shift according to the meeting format or topic, allowing realistic group dynamics and participant evolution.  
An LLM also checks for contradictory traits (e.g., proactive yet blocking) to maintain role consistency.

\begin{figure}[t]
    \centering
    \includegraphics[width=0.9\linewidth]{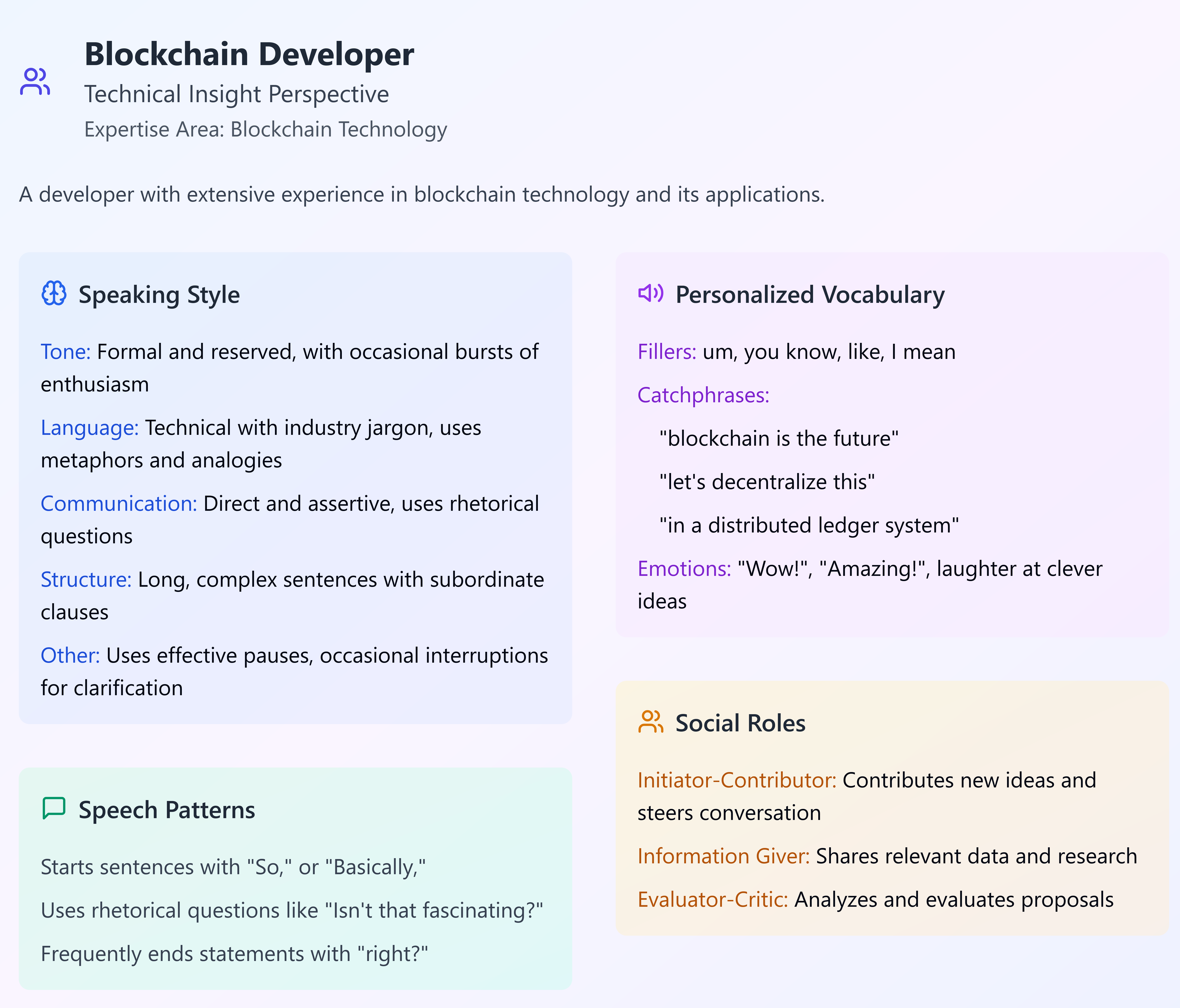}
    \caption{Example of a minimal participant definition.}
    \label{fig:participant_figure}
\end{figure}

\paragraph{Stage 3: Scripting.}
An LLM expands the target summary into a flexible outline of thematic scenes, each dedicated to a major subtopic of the knowledge source.
A scene includes a title, an agenda description, and bullet points covering key content.
Scenes also specify relevant aspects of the source material and invite participants to draw on personal experiences, reflecting each persona.
Depending on the meeting type, extra scenes (e.g., a pre-meeting brainstorming phase) may be added for realism, even if they are not part of the target summary.
This outline serves as a roadmap, allowing participants to briefly shift themes or discuss subtopics without straying from the broader structure.

\subsection{Production Phase}
This phase simulates turn-by-turn dialogue among multiple LLMs and validate each scene for quality.

\paragraph{Stage 4: Filming.}
\label{sec:stage_4_filming}

We generate transcripts scene by scene, following the outline from Stage 3: Scripting.
Each participant is an independent LLM instance, contributing one turn at a time.
To emulate real-world scenarios \cite{ZhouSEK24}, we implement a non-omniscient approach where each participant has private memory, seeing only their own profile, relevant source snippets, a summary of prior scenes, the last three turns of the previous scene and all turns so far in the current scene.

Scenes begin with the participant most relevant to the topic, determined by the exact LLM-based matching used during casting \cite{LiWXG25}.

At the end of each turn, the current speaker nominates the next speaker based on roles, current focus, and prior contributions (e.g., previously raised concerns or ideas introduced) \cite{NonomuraM24a}.
Reminding the current speaker about the other participants prevents them from being left out.

To conclude a turn, the speaker can propose a vote to end a scene \cite{WangWST24a}.
The vote can be called after the first turn if the speaker finds that they have stated all the details in their individual knowledge that fit the current discussion topic.
If the majority (more than 50\%) of the participants agree during voting, the scene ends.
To avoid endless loops, a system reminds the group to finalize if no one initiates a vote after 50 turns.
\pipeline{} then proceeds to the next scene or terminates if all scenes are complete.

\paragraph{Stage 5: Quality assuring.}
Inspired by 'Self-Refinement' \cite{MadaanTGH23b} where LLMs iteratively provide feedback and refine their output, we use a ``director'' model to review each scene along three dimensions: \textbf{topical alignment}, i.e., adherence to the target summary \cite{LinC23}, \textbf{conversational naturalness}, i.e., turn-taking quality, dialogue flow \cite{LiuIXW23}, and \textbf{coherence/factual accuracy}, i.e., logical progression, consistency with knowledge source \cite{XieZPJ24}.
The director LLM provides feedback to correct any critical issues identified (e.g., missing subtopics, overly formal language, and contradictions).
We allow up to three ``re-filming'' cycles, considering this feedback.
If a scene remains problematic after three attempts, \pipeline{} proceeds with the best version, deferring residual edits to the post-production phase.
In practice, this fallback was never triggered for \dataset{}.

\subsection{Post-Production Phase}
This phase injects disruptions for realism and polishes transcripts to ensure real-sounding meetings.

\paragraph{Stage 6: Special effects.}
We inject flow-breaking events (e.g., phone calls, technical glitches, side questions) with a 25\% chance per scene\footnote{Chosen empirically to balance realism without overwhelming the meeting.}, permitting multiple disruptions per meeting.
These events lead to adding a few turns from the participants (e.g., acknowledging a ringing phone) before the main discussion resumes.

\paragraph{Stage 7: Editing.}
A two-step linguistic refinement removes repetitive phrases and overly formal speech and adds minor speech markers (e.g., hesitations).
A subsequent detector–revision step targets any remaining synthetic cues (e.g., uniformly polite turns, unrealistic consensus).
If necessary, minor disagreements or paraphrases are introduced to achieve more natural, human-like transcripts.

\begin{table*}[t]
    \renewcommand{\arraystretch}{1.2} 
    \tiny
    \centering
    \begin{tabular}{lllllllllll}
        \toprule
        \rowcolor{gray!20} 
        Dataset & \# Meetings & \# Speaker & \# Unique Spea. & \# Turns & \# Words & Vocab. & Token Overlap & Sum. Len. & Interruptions & Language \\
        \midrule
        AMI   & 137 & $4.0_{\textit{0.00}}$ & 4    & $513.5_{\textit{266.2}}$ & $4937.5_{\textit{1999.3}}$ & 9388 & - & $109.9_{\textit{27.1}}$ & no &  informal \\
        ICSI  & 44  & $6.2_{\textit{1.3}}$   & 35   & $757.5_{\textit{374.8}}$ & $9889.4_{\textit{3794.9}}$ & 9164 & - & $93.3_{\textit{22.2}}$  & no &  formal \\
        WCPC  & 51  & $16.8_{\textit{18.7}}$ & 316  & $337.3_{\textit{277.3}}$ & $11427.8_{\textit{4574.0}}$ & 13780 & - & $122.3_{\textit{39.2}}$  & no &  informal \\
        QMSum & 232 & $7.2_{\textit{10.1}}$  & 330  & $521.0_{\textit{320.4}}$ & $7303.4_{\textit{4232.2}}$  & 20505 & - & $109.5_{\textit{30.7}}$  & no &  both \\
        \midrule
        EN    & 500 & $5.1_{\textit{2.8}}$  & 3200 & $405.0_{\textit{330.3}}$ & $6223.4_{\textit{4084.4}}$ & 10347 & 0.081 & $207.7_{\textit{22.7}}$ & yes ($\sim$ 0.5)&  both  \\
        GER   & 300 & $5.0_{\textit{2.8}}$  & 1000 & $393.3_{\textit{323.2}}$ & $6272.4_{\textit{3793.2}}$ & 9589  & 0.096 & $170.3_{\textit{29.0}}$ & yes ($\sim$ 0.5) & both  \\
        \bottomrule
    \end{tabular}
    \caption{Statistics on \dataset{} for English (EN) and German (GER) and established corpora. Values are Mean$_{Std}$.}
    \label{tab:corpora_metrics}
\end{table*}

\section{The \dataset{} Dataset}
\label{sec:data_insights}

In the following, we detail how \dataset{} was generated, present its overall statistics, and analyze the authenticity of its simulated meetings and how challenging they are compared to real transcripts.
All data and annotations performed here are available per \Cref{sec:Data_Availability}.

\subsection{Setup to generate \dataset{}}

\paragraph{Knowledge source.} 
Unlike synthetic datasets built on minimal or artificial context, \dataset{} builds on grounded text from Wikipedia.
We select 300 articles from 28 broad domains (e.g., Global Issues, Technological Innovations, Cultural Diversity, Philosophy, Environment \& Ecology), each meeting three criteria: (1) have at least five subsections, (2) no negative flags regarding article quality, and (3) no reference deficiencies (to avoid contradictory claims).
For the German subset, we choose 150 German-language articles from the same pool, ensuring a high content overlap with their English counterparts via BERTScore \cite{ZhangKWW20} and an empirically chosen threshold of 0.7.
For each article, we randomly assign meeting types, participant roles, and the number of participants, yielding 500 English and 300 German meetings.

\paragraph{Backbone model.}
We use GPT \cite{OpenAIAAA24} for all stages of \pipeline{}, taking advantage of its 128k-token context window and robust role-playing capabilities \cite{KirsteinRKG24a}.
In \Cref{sec:ablation}, we show that models with fewer parameters (Gemini, DeepSeek, Llama) can generate high-quality transcripts with minor drops in naturalness.

\paragraph{Baseline.}
We compare \dataset{} to QMSum \cite{ZhongYYZ21h}, an established dataset combining academic (ICSI \cite{JaninBEE03}), product (AMI \cite{MccowanCKA05}), and parliament (Welsh/Canadian, WPCP) meetings.
While other corpora exist (e.g., MeetingBank \cite{HuGDD23a}, ELITR \cite{NedoluzhkoSHG22}), they closely resemble QMSum’s formal institutional settings.
We also contrast our non-omniscient multi-LLM pipeline with a single omniscient GPT approach \cite{ZhouSEK24} that generates entire meetings in one shot given the same article, target summary, speaker count, and meeting type used for \dataset{}.
To match in meeting length, we prompt GPT multiple times to circumvent its 16k-token output limit and produce transcripts of $\sim$20k tokens.

\subsection{General Statistics}
\noindent
\textbf{Subsection key finding.}
\dataset{}’s diversity enables large-scale benchmarking of summarizers, mimicking the varied conditions seen in real meetings.

\paragraph{Comparison.}

\Cref{tab:corpora_metrics} summarizes statistics for \dataset{}.
Our average meeting length ($\sim$6,200 words), turn count ($\sim$400 turns), and participant count ($\sim$4) closely mirrors real corpora (e.g., AMI), though \dataset{} has higher variance (e.g., larger standard deviation for turn counts).

\paragraph{Unique features.}
In addition to standard business, academic, and parliamentary contexts, \dataset{} spans 14 new meeting types and 28 Wikipedia-based domains.
\dataset{}’s dialogues primarily rephrase rather than copy source passages (average token overlap of English: 0.081, German: 0.096).
Over 3,000 participanth adopt unique speech patterns and undergo up to four behavioral shifts, introducing dynamic role changes beyond the fixed-character setups typical of existing corpora.
One-third of the meetings have interruptions, with 20\% of those featuring three or more disruptions. 
Existing corpora typically lack these unpredictabilities.

\subsection{Authenticity Evaluation}
\label{sec:authenticity_eval}

\noindent
\textbf{Subsection key finding.}
\dataset{} meetings exhibit near-real conversation flow, and participants behave in ways that closely match human expectations.

\paragraph{Approach.}

We evaluate meeting authenticity along two axes: overall authenticity, i.e., coherence, consistency, interestingness, naturalness \cite{ChenPTK23}, and participant behavior authenticity, e.g., conflicts and power dynamics, defined in consultation with psychology and sociology experts.
Therefore, eight overarching behavior categories, i.e., knowledge, power, conflict, status, trust, support, similarity, and fun \cite{bales2009interaction,ChoiAVQ20}, were divided into 18 items complete list given in \Cref{tab:appendix_behavior_questions} in \Cref{sec:appendix_E}).

Six annotators (students to PhD candidates in Psychology, Computer Science, Communication; aged 23–28, at least C1 English/German) rate 30 English and 30 German meetings from \dataset{}, as well as 30 meetings from QMSum, using a 1–5 Likert scales Krippendorff's $\alpha = 0.83$\footnote{
To support broader community use, we extend annotations across the entire \dataset{}, we used GPT with the three-step approach of \citet{KirsteinLG25}, aligned automatic scores through the self-training concept to match human ratings until the average discrepancy was below 0.8 points.
}.
Because existing real-meeting corpora may exhibit a behavior bias due to the formal or staged nature, we also surveyed 100 crowdworkers (age 24–63, balanced by gender, diverse professional backgrounds such as law, engineering, and management) who frequently attend meetings to collect experiences with real meetings.
Additional details on annotators, crowdworkers, and reliability are in \Cref{sec:appendix_F}.

\begin{table}[t]
    \centering
    \renewcommand{\arraystretch}{1.2} 
    \scriptsize
    \setlength{\tabcolsep}{3.8pt} 

    \definecolor{highlightGreen}{HTML}{D4F4E9} 

    \begin{tabular}{lccccccc}
        \toprule
        \rowcolor{gray!20} 
         & \textbf{EN} & \textbf{GER} & \textbf{AMI} & \textbf{ICSI} & \textbf{WPCP} & \textbf{QMSum} & \textbf{OMNI}\\
        \midrule
        COH 
            & \cellcolor{highlightGreen}{$\textbf{4.5}_{\textit{0.00}}$}
            & \cellcolor{highlightGreen}{$\textbf{4.5}_{\textit{0.18}}$}
            & \cellcolor{highlightGreen}{$\textbf{4.5}_{\textit{0.36}}$}
            & $3.5_{\textit{0.36}}$
            & \cellcolor{highlightGreen}{$\textbf{4.5}_{\textit{0.00}}$}
            & \cellcolor{highlightGreen}{$\textbf{4.5}_{\textit{0.73}}$}
            & $3.5_{\textit{0.00}}$\\[1ex]
        CON 
            & \cellcolor{highlightGreen}{$\textbf{4.5}_{\textit{0.07}}$}
            & \cellcolor{highlightGreen}{$\textbf{4.5}_{\textit{0.09}}$}
            & \cellcolor{highlightGreen}{$\textbf{4.5}_{\textit{0.68}}$}
            & $3.5_{\textit{0.87}}$
            & \cellcolor{highlightGreen}{$\textbf{4.5}_{\textit{0.38}}$}
            & $4_{\textit{0.59}}$
            & $3.5_{\textit{0.09}}$\\[1ex]
        INT 
            & \cellcolor{highlightGreen}{$\textbf{4.5}_{\textit{0.13}}$}
            & \cellcolor{highlightGreen}{$\textbf{4.5}_{\textit{0.23}}$}
            & $4_{\textit{0.68}}$
            & $3_{\textit{0.87}}$
            & $4_{\textit{0.38}}$
            & $4_{\textit{0.77}}$
            & $4_{\textit{3.08}}$\\[1ex]
        NAT 
            & $4.5_{\textit{0.12}}$
            & $4_{\textit{1.37}}$
            & $4.5_{\textit{0.55}}$
            & $4.5_{\textit{0.82}}$
            & \cellcolor{highlightGreen}{$\textbf{5}_{\textit{0.00}}$}
            & $4.5_{\textit{0.90}}$
            & $3_{\textit{0.74}}$\\
        \bottomrule
    \end{tabular}
    \caption{Evaluation of synthetic meetings on COH = coherence, CON = consistency, INT = interestingness, and NAT = naturalness. Values are Median$_{Std}$. EN, GER are \dataset{} English and German. OMNI is a single LLM. \textbf{Higher} is better.}
    \label{tab:common_dimensions_evaluation}
\end{table}

\paragraph{Quantitative analysis.}
\Cref{tab:common_dimensions_evaluation} compares overall authenticity scores for \dataset{} and QMSum.
The English/German \dataset{} subsets match QMSum in coherence, while both subsets score higher in conciseness and interestingness (4.5/5 in \dataset{} vs.\ 4/5 in QMSum). 
ICSI meetings receive the lowest interestingness rating (3/5) due to slower topic evolution, whereas WPCP leads in naturalness (5/5). 
The German \dataset{} subset lags the English one by 0.5 points in naturalness (English: 4.5/5), which is similar to most QMSum data.
Notably, transcripts generated by the single omniscient model rank lowest overall due to shallow, repetitive content.

\begin{figure}[t]
    \centering
    \includegraphics[width=0.8\linewidth]{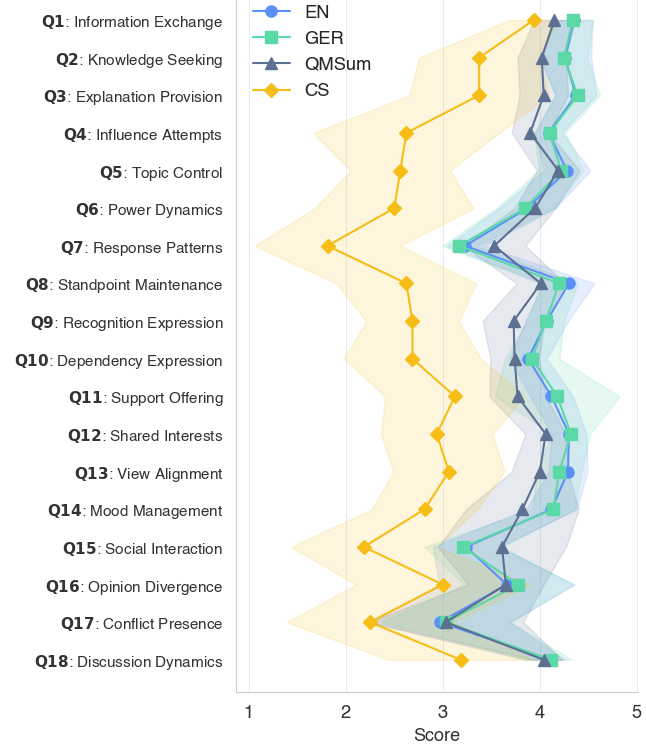}
    \caption{Evaluation of the social behavior in the meetings. All questions are detailed in \Cref{sec:appendix_A}. EN and GER are the English and German subsets of \dataset{}, and CS indicates the crowdsourced experiences.}
    \label{fig:psychology_grounded_evaluation}
\end{figure}

As shown in \Cref{fig:psychology_grounded_evaluation}, social behavior patterns in \dataset{} align closely with both QMSum (within 0.2 points) and the crowdsourced baseline (within 1.0 points). 
Modest differences likely reflect gaps between self-assessment and observed performance \cite{kruger1999unskilled},  often linked to social desirability or recency biases.
The synthetic meetings’ mild overperformance in some categories suggests a slight dramatization of behaviors.
We show the behavior pattern in single-LLM meetings in \Cref{sec:appendix_A}, which just minimal overlaps with the \dataset{} patterns.
In sum, \pipeline{} speakers reliably replicate real behaviors.

\paragraph{Qualitative insights.}
Annotators highlight the realistic interplay of participant knowledge, consistent role portrayal, and dynamic sub-topic exploration.
Occasionally, overuse of transitional phrases (e.g., “This sounds great, but...”) or overly cordial tones reveal the synthetic origin (noted in 57 out of 800 meetings).
Despite these artifacts, \dataset{} remains a near-realistic meeting proxy, offering a robust environment to develop and benchmark summarization methods.
In contrast, the omniscient approach often yields shallow dialogues, word/passage repetition, or even recitations from well-known sources (e.g., the Bible).

\subsection{Transcript Challenge Assessment}
\label{sec:challenges}
\noindent
\textbf{Subsection key finding.} 
\dataset{} preserves speaker-related complexities and extends information-centric challenges closer to real meetings.

\paragraph{Analysis.}
We adopt the seven challenges from \citet{KirsteinWRG24a}, i.e., spoken language, speaker dynamics, coreference, discourse structure, contextual turn-taking, implicit context, and low information density, to gauge the difficulty of summarizing \dataset{} transcripts (\Cref{tab:challenge_scores}).
We compare \dataset{}’s median 1–5 Likert ratings against human-annotated QMSum data \cite{KirsteinWRG24a}.
Overall, \dataset{} matches QMSum in spoken language (3/5) and coreference (2/5, German subset increases to 3/5), though it exhibits calmer speaker dynamics (2/5 vs.\ $\sim$3/5 in QMSum).
Both \dataset{} subsets surpass QMSum in implicit context (\dataset{}: 4/5 vs.\ QMSum: 0/5) and low information density (\dataset{}: 4/5 vs.\ QMSum: 2.5/5), indicating that participants rely on prior exchanges instead of explicitly reiterating every detail.
By contrast, many existing corpora emphasize exhaustive information sharing (ICSI, WPCP) or feature staged, side-discussion-free scenarios (AMI).

\definecolor{highlightGreen}{HTML}{D4F4E9} 

\begin{table}[t]
    \centering
    \renewcommand{\arraystretch}{1.2} 
    \scriptsize
    \setlength{\tabcolsep}{4pt}
    \begin{tabular}{l c c c c c c}
        \toprule
        \rowcolor{gray!20} 
        \textbf{Challenge} & \textbf{EN} & \textbf{GER} & \textbf{AMI} & \textbf{ICSI} & \textbf{WCPC} & \textbf{QMSum} \\
        \midrule
        spoken language 
            & $3_{\textit{0.49}}$
            & $3_{\textit{0.54}}$            
            & \cellcolor{highlightGreen}{$\textbf{4}_{\textit{0.22}}$}
            & $3_{\textit{0.70}}$
            & $0.5_{\textit{0.55}}$
            & \cellcolor{highlightGreen}{$\textbf{3}_{\textit{1.35}}$} \\           
        speaker dynamics 
            & $2_{\textit{0.62}}$
            & $2_{\textit{0.73}}$            
            & \cellcolor{highlightGreen}{$\textbf{4}_{\textit{0.66}}$}
            & $3_{\textit{0.70}}$
            & $2_{\textit{0.63}}$
            & $3_{\textit{0.91}}$ \\           
        coreference 
            & $2_{\textit{0.74}}$
            & \cellcolor{highlightGreen}{$\textbf{3}_{\textit{0.80}}$}            
            & $2_{\textit{0.76}}$
            & $2_{\textit{1.07}}$
            & $1.5_{\textit{0.55}}$
            & $2_{\textit{0.90}}$ \\
        discourse structure
            & \cellcolor{highlightGreen}{$\textbf{3}_{\textit{1.00}}$}
            & $2_{\textit{1.20}}$
            & \cellcolor{highlightGreen}{$\textbf{3}_{\textit{0.93}}$}
            & $2_{\textit{0.84}}$
            & \cellcolor{highlightGreen}{$\textbf{3}_{\textit{1.17}}$}
            & \cellcolor{highlightGreen}{$\textbf{3}_{\textit{0.96}}$} \\           
        turn-taking 
            & \cellcolor{highlightGreen}{$\textbf{4}_{\textit{0.56}}$}
            & $3.5_{\textit{0.51}}$
            & $3_{\textit{0.93}}$
            & $3_{\textit{0.67}}$
            & $2_{\textit{0.84}}$
            & $3_{\textit{1.04}}$ \\           
        implicit context 
            & \cellcolor{highlightGreen}{$\textbf{4}_{\textit{0.16}}$}
            & \cellcolor{highlightGreen}{$\textbf{4}_{\textit{0.18}}$}
            & $0_{\textit{0.00}}$
            & $2_{\textit{0.97}}$
            & $0_{\textit{0.82}}$
            & $0_{\textit{0.85}}$ \\       
        information density
            & \cellcolor{highlightGreen}{$\textbf{4}_{\textit{0.27}}$}
            & \cellcolor{highlightGreen}{$\textbf{4}_{\textit{0.00}}$}
            & $3_{\textit{0.55}}$
            & $2_{\textit{0.57}}$
            & $2_{\textit{0.84}}$
            & $2.5_{\textit{0.88}}$\\            
        \bottomrule
    \end{tabular}
    \caption{Scores on challenges of meeting transcripts. EN, GER are \dataset{} English and German. 
    Values are Median$_{Std}$. 
    \textbf{Higher} is more difficult.}
    \label{tab:challenge_scores}
\end{table}

\section{Baseline Models and Results}
\label{sec:experiments}

\begin{table*}[t]
    \centering
    \renewcommand{\arraystretch}{1.2} 
    \scriptsize
    \setlength{\tabcolsep}{2.5pt} 

    \definecolor{highlightGreen}{HTML}{D4F4E9}

    \begin{tabular}{@{}l@{\hspace{2pt}}cccc c cccc c cccc@{}}
        \toprule
        
            & \multicolumn{4}{c}{\textbf{QMSum}} 
            & \multicolumn{1}{c}{}
            & \multicolumn{4}{c}{\textbf{\dataset{} [EN]}}
            & \multicolumn{1}{c}{}
            & \multicolumn{4}{c}{\textbf{\dataset{} [GER]}} \\
        \cmidrule(lr){2-5} \cmidrule(lr){7-10} \cmidrule(lr){12-15}
            & GPT & Gemini & Deepseek & Llama 
            &  & GPT & Gemini & Deepseek & Llama 
            &  & GPT & Gemini & Deepseek & Llama \\
        \midrule
        \rowcolor{gray!20} 
        \multicolumn{15}{c}{\textbf{Meeting Summary Focused Evaluation Metric} (lower is better)} \\
        \midrule
        Coreference
            & \cellcolor{highlightGreen}$0_{\textit{1.22}}$ 
            & $3_{\textit{1.58}}$
            & \cellcolor{highlightGreen}$0_{\textit{1.42}}$
            & $1_{\textit{1.67}}$
            & 
            & \cellcolor{highlightGreen}$0_{\textit{1.45}}$
            & $3_{\textit{1.57}}$
            & $1_{\textit{1.54}}$
            & $2_{\textit{1.60}}$
            &
            & \cellcolor{highlightGreen}$0_{\textit{1.38}}$
            & $3_{\textit{1.57}}$
            & $1.5_{\textit{1.68}}$
            & $2_{\textit{1.52}}$ \\
        Hallucination     
            & $3_{\textit{1.22}}$
            & $4_{\textit{2.04}}$
            & \cellcolor{highlightGreen}$2_{\textit{1.88}}$
            & $3_{\textit{2.08}}$
            & 
            & $4_{\textit{0.98}}$
            & $4_{\textit{1.40}}$
            & $4_{\textit{1.81}}$
            & $4_{\textit{1.02}}$
            &
            & $4_{\textit{1.57}}$
            & $4_{\textit{1.65}}$
            & \cellcolor{highlightGreen}$3_{\textit{1.81}}$
            & $4_{\textit{1.61}}$ \\
        Incoherence     
            & $4_{\textit{1.50}}$
            & $4_{\textit{1.09}}$
            & $4_{\textit{1.85}}$
            & $4_{\textit{1.88}}$
            & 
            & $4_{\textit{0.94}}$
            & $4_{\textit{0.72}}$
            & $4_{\textit{0.94}}$
            & $4_{\textit{1.43}}$
            &
            & $4_{\textit{1.18}}$
            & $4_{\textit{1.39}}$
            & \cellcolor{highlightGreen}$3_{\textit{1.20}}$
            & \cellcolor{highlightGreen}$3_{\textit{1.57}}$ \\
        Irrelevance     
            & \cellcolor{highlightGreen}$2_{\textit{1.70}}$
            & $3_{\textit{1.32}}$
            & \cellcolor{highlightGreen}$2_{\textit{1.48}}$
            & \cellcolor{highlightGreen}$2_{\textit{1.52}}$
            & 
            & $3_{\textit{1.14}}$
            & $3_{\textit{1.07}}$
            & $3_{\textit{1.11}}$
            & \cellcolor{highlightGreen}$2_{\textit{1.42}}$
            &
            & $2_{\textit{1.16}}$
            & $2_{\textit{1.23}}$
            & $2_{\textit{1.29}}$
            & $2_{\textit{1.27}}$ \\
        Language     
            & \cellcolor{highlightGreen}$1_{\textit{1.30}}$
            & $2_{\textit{1.44}}$
            & $2_{\textit{1.44}}$
            & \cellcolor{highlightGreen}$1_{\textit{1.50}}$
            & 
            & $1_{\textit{1.17}}$
            & $1_{\textit{1.20}}$
            & $1_{\textit{1.22}}$
            & $1_{\textit{1.04}}$
            &
            & \cellcolor{highlightGreen}$0_{\textit{1.31}}$
            & \cellcolor{highlightGreen}$0_{\textit{0.97}}$
            & \cellcolor{highlightGreen}$0_{\textit{1.04}}$
            & $1_{\textit{1.31}}$ \\
        Omission     
            & $3_{\textit{0.40}}$
            & $3_{\textit{0.38}}$
            & $3_{\textit{0.36}}$
            & $3_{\textit{0.47}}$
            & 
            & $4_{\textit{0.16}}$
            & $4_{\textit{0.31}}$
            & $4_{\textit{0.48}}$
            & $4_{\textit{0.31}}$
            &
            & $4_{\textit{0.18}}$
            & $4_{\textit{0.00}}$
            & $4_{\textit{0.18}}$
            & $4_{\textit{0.18}}$ \\
        Repetition     
            & $4_{\textit{1.05}}$
            & \cellcolor{highlightGreen}$3_{\textit{0.98}}$
            & \cellcolor{highlightGreen}$3_{\textit{1.01}}$
            & $4_{\textit{1.19}}$
            & 
            & $4_{\textit{0.74}}$
            & $4_{\textit{0.44}}$
            & $4_{\textit{0.39}}$
            & $4_{\textit{0.51}}$
            &
            & $4_{\textit{0.52}}$
            & $4_{\textit{0.79}}$
            & $4_{\textit{0.34}}$
            & $4_{\textit{0.47}}$ \\
        Structure     
            & $4_{\textit{0.90}}$
            & \cellcolor{highlightGreen}$3_{\textit{1.70}}$
            & \cellcolor{highlightGreen}$3_{\textit{1.65}}$
            & $4_{\textit{1.69}}$
            & 
            & $3_{\textit{1.57}}$
            & $3_{\textit{1.53}}$
            & $3_{\textit{1.52}}$
            & $3_{\textit{1.36}}$
            &
            & $3_{\textit{1.66}}$
            & $3_{\textit{1.42}}$
            & $3_{\textit{1.50}}$
            & $3_{\textit{1.54}}$ \\
        \midrule
        \rowcolor{gray!20} 
        \multicolumn{15}{c}{\textbf{General Evaluation Metrics} (higher is better)} \\
        \midrule
        R-1
            & $37.73_{\textit{5.85}}$
            & $39.61_{\textit{7.21}}$
            & $32.37_{\textit{5.18}}$
            & \cellcolor{highlightGreen}$40.91_{\textit{5.21}}$
            & 
            & $39.68_{\textit{5.73}}$
            & $38.82_{\textit{5.79}}$
            & $33.18_{\textit{7.38}}$
            & \cellcolor{highlightGreen}$40.35_{\textit{5.46}}$
            &
            & $33.74_{\textit{4.01}}$
            & $32.03_{\textit{5.14}}$
            & \cellcolor{highlightGreen}$33.90_{\textit{5.37}}$
            & $32.54_{\textit{4.50}}$ \\
        R-2
            & $7.95_{\textit{4.18}}$
            & \cellcolor{highlightGreen}$11.10_{\textit{4.86}}$
            & $5.80_{\textit{2.94}}$
            & $9.79_{\textit{4.08}}$
            & 
            & $8.43_{\textit{3.45}}$
            & \cellcolor{highlightGreen}$8.96_{\textit{3.50}}$
            & $8.07_{\textit{3.33}}$
            & $8.62_{\textit{3.18}}$
            &
            & $7.45_{\textit{2.98}}$
            & $6.98_{\textit{3.72}}$
            & \cellcolor{highlightGreen}$7.93_{\textit{3.67}}$
            & $7.24_{\textit{3.17}}$ \\
        R-L
            & $21.39_{\textit{4.05}}$
            & \cellcolor{highlightGreen}$27.55_{\textit{6.36}}$
            & $18.46_{\textit{3.44}}$
            & $7.95_{\textit{4.18}}$
            & 
            & \cellcolor{highlightGreen}$29.98_{\textit{4.88}}$
            & $27.81_{\textit{4.18}}$
            & $23.16_{\textit{5.98}}$
            & $29.45_{\textit{3.67}}$
            &
            & \cellcolor{highlightGreen}$25.20_{\textit{3.68}}$
            & $23.59_{\textit{4.32}}$
            & $24.60_{\textit{3.29}}$
            & $24.10_{\textit{3.11}}$ \\
        BS (F1)
            & $61.61_{\textit{2.87}}$
            & $60.64_{\textit{3.66}}$
            & $59.44_{\textit{2.78}}$
            & \cellcolor{highlightGreen}$62.28_{\textit{2.76}}$
            & 
            & \cellcolor{highlightGreen}$63.80_{\textit{3.11}}$
            & $63.66_{\textit{2.49}}$
            & $63.16_{\textit{3.37}}$
            & $63.53_{\textit{2.72}}$
            &
            & $64.75_{\textit{1.20}}$
            & $64.58_{\textit{1.74}}$
            & \cellcolor{highlightGreen}$64.88_{\textit{1.77}}$
            & $64.61_{\textit{1.33}}$ \\
        \bottomrule
    \end{tabular}
    \caption{Combined Evaluation Results: Human Evaluation top and Automatic Evaluation bottom. 
    Values are Median$_{Std}$. MESA scores are 1--5 Likert ratings, 
    ROUGE (R-1/R-2/R-L) and BERTScore (BS) are 0--100.}
    \label{tab:summary_performance}
\end{table*}

We evaluate current LLMs on abstractive meeting summarization for real and synthetic transcripts, sampling 30 meetings from each \dataset{} (English, German) and QMSum (90 total).

\subsection{Experimental Setup}
\paragraph{Summarization approaches.} 
We benchmark two closed-weight models (GPT, Gemini) and two open-weight models (Llama, DeepSeek), excluding refinement-based methods \cite{KirsteinLG25a}, as these would produce self-refined GPT summaries.
We use a simple zero-shot prompt requesting an abstract summary of up to 250 tokens, reflecting standard practices in meeting summarization \cite{KirsteinLG25a}.
Full prompt details are provided in \Cref{sec:appendix_A}.

\paragraph{Evaluation metrics.}
We compare system outputs against reference summaries using the established ROUGE (R-1/R-2/R-L) \cite{Lin04} and BERTScore (rescaled F1) \cite{ZhangKWW20} metrics along MESA \cite{KirsteinLG25}, an LLM-based metric for the error types of meeting summarization (e.g., structure, irrelevance).
These metrics enable direct comparisons with prior work and provide detailed insights into model weaknesses.

\subsection{Results}
\label{subsec:results_summ}

\Cref{tab:summary_performance} contains the evaluation scores.

\paragraph{Reasoning boosts performance.}
Although Llama tops the ROUGE/BERTScore metrics on both QMSum and \dataset{}, MESA shows that DeepSeek consistently matches or improves over other models by minimizing common error types ($\sim$1 point lower per category), especially on QMSum.
DeepSeek also outperforms on the German \dataset{} subset, while Gemini generally trails behind.
All models perform comparably on the English subset, with the least language and coreference issues ($\sim$ 1/5).

\paragraph{\dataset{} reveals LLM struggles.} 
MESA scores for categories like incoherence, structure, and repetition remain similar to those on QMSum ($\sim$ 4/5) but with lower deviation for the \dataset{} subsets.
We conclude that the varied topics and meeting formats in \dataset{} add to the overall difficulty and negatively influence summary quality.

\paragraph{Contextualization deficits persist.}
The omission and irrelevance rows show that all models struggle with \dataset{}'s more difficult cross-turn information and information scarcity observed in \Cref{sec:challenges}.
Omission rises from 3/5 on QMSum to 4/5 on \dataset{}, and irrelevance from 2/5 to 3/5 for the English subset (German remains 2/5).
We derive that reliable content understanding \cite{KirsteinRKG24a,KirsteinLG25a} can be tested with our \dataset{} and that current LLMs struggle with this.

\section{Ablations on \pipeline{}}
\label{sec:ablation}
Our ablations analyze \pipeline{} under different settings and assess its role consistency.
Below, we briefly summarize the experiments; full details, tables, and figures appear in \Cref{sec:appendix_H}.
All evaluations use the same model settings, metrics, and annotation guidelines from \Cref{sec:data_insights,sec:experiments}.

\paragraph{Knowledge source shapes discussion depth and structure.} 
Beyond the semi-structured Wikipedia articles used for \dataset{}, we test \pipeline{} on 10 research papers from arXiv \cite{CohanDKB18} (clearly sectioned, e.g., abstract, methods, results), 10 human-written stories from WritingPrompts \cite{FanLD18a} (unstructured) and 10 research papers from PubMed \cite{missing} (clearly structured, niche domain knowledge).
Using GPT as the backbone, we generate one synthetic meeting per source (30 total) and apply our evaluation framework (\Cref{sec:data_insights}).
As shown in \Cref{tab:app_knowledge_source} (overall authenticity) and \Cref{fig:app_psychology1} (behavior authenticity) in \Cref{sec:app_knowledge_source}, quantitative metrics remain stable, suggesting meeting naturalness is not strongly tied to input structure. 
Research papers lead to deeper discussions, while short stories produce briefer, shallow meetings.

\paragraph{Other models yield shorter meetings.} 
To see if other LLMs can drive \pipeline{}, we replace GPT with DeepSeek and Llama, generating 25 meetings per model using the \dataset{} Wikipedia article pool.
We assess quality (overall authenticity, behavior authenticity, challenges) and compare outputs using identical knowledge sources.
The results are given in \Cref{tab:app_mid_size_models} (overall authenticity) and \Cref{fig:app_psychology2} (behavior authenticity) in \Cref{sec:app_mid_size_models}.
Although these models produce transcripts with about 50 fewer turns than \dataset{}, their naturalness remains high (4/5), and they replicate participant roles and behaviors almost as consistently as GPT, close to human expectations.
Occasionally, transcripts feature fewer back-and-forth exchanges but remain high-quality in qualitative reviews, thereby broadening \pipeline{}’s real-world applicability.

\paragraph{Editing is model-dependent but rarely critical.}
Stage 7 (editing) addresses issues such as formal phrasing or repeated filler words.
To measure its influence, we review 75 transcripts (25 each from GPT, DeepSeek, and Llama), examining chain-of-thought logs of stage 7 and final transcripts.
Only \emph{1 in 25} GPT transcripts require major edits to mask synthetic traits, rising to \emph{2 in 25} for the Llama-based models\footnote{\dataset{} contains a flag if refinement is due to a pipeline issue or a model-specific artifact along the feedback from the editorial stage.}. 
Minor refinements correct model-specific wording or repeated transitions, indicating that \pipeline{} already yields coherent discussions while stage 7 polishes for full realism.

\paragraph{Roles and behaviors are reliably enacted.}
Drawing on \citet{SerapioGarciaSCS23a}, we evaluate whether participants preserve assigned roles and behaviors.
We select 30 participants from the newly generated meetings (10 from GPT, Gemini, DeepSeek, and Llama) and ask our six annotators to judge each turn’s alignment with predefined behavior (e.g., blocker).
We cover 100 scenes and 400 simulated participants in total.
More than 95\% of GPT turns match their roles, dropping slightly to 92\% for the other two\footnote{Annotation will be extended and made available later.}.
To cross-check, we sample 50 participant profiles and prompt GPT enacting these profiles to answer the TREO questionnaire \cite{MathieuTKD15} containing 48 1--5 Likert-scored questions to determine someone's role in a group.
We find 48 of the profiles consistent with the assigned behaviors\footnote{All responses are provided as per \Cref{sec:Data_Availability}.}.
We conclude that \pipeline{} enforces coherent persona dynamics for social simulation \cite{ZhouZMZ24}.

\paragraph{Fine-tuning on \dataset{} improves QMSum performance.}
Fine-tuning \textsc{Llama-3.2-3B} on 100 synthetic \dataset{} meetings and evaluating on the 35-meeting \textsc{QMSum} test split produces clear transfer gains.  
Compared with the non-fine-tuned model, fine-tuning on \dataset{} lowers \textsc{MESA} errors, and, against an equally sized QMSum fine-tuned checkpoint, further reduces \emph{coreference}, \emph{incoherence}, and \emph{language} errors while trailing slightly in \emph{irrelevance} and \emph{structure}.  
ROUGE and BERTScore differ by no more than two points between the two fine-tuned variants, and manual inspection confirms that fine-tuning on \dataset{} yields the most coherent, information-dense summaries.  
These results position \dataset{} as a practical proxy when only limited labelled meeting data are available.


\section{Final Considerations}
We introduced \pipeline{}, a seven-stage multi-agent framework that uses psychologically grounded, non-omniscient LLMs to generate source-grounded meeting transcripts.
\pipeline{} generated \dataset{}, a multilingual corpus of 500 English and 300 German meetings on 28 domains and 14 meeting types.
Human assessments showed that \dataset{} closely mirrors real meetings (4.5/5 in naturalness) and amplified low-information density (4/5).
Comparisons with real meetings suggest that \dataset{} captures authentic group dynamics, while our ablation studies highlighted how varying knowledge sources and backbone models shape transcript quality and demonstrate the GPT's reliable role enactment.

\dataset{} and its detailed annotations open new directions for meeting summarization, from multilingual model development to re-introducing fine-tuning for LLMs and reinforcement learning to address persistent shortcomings (\Cref{sec:experiments}).
By releasing \pipeline{} as open-source, we provide a powerful toolkit that researchers can adapt to low-resource languages, diverse domains, and a range of conversational styles. 
The framework’s psychology-based behavior definitions and evaluation methodology bring a higher level of realism into synthetic conversations, enabling deeper investigations of social dynamics.
Our work bridges a data gap through human-like meeting simulations that foster advances in summarization, conversational AI, social simulation, and beyond.

\section*{Limitations}
The quality of generated meetings partly depends on the underlying LLM's capabilities and biases.
Models with smaller context windows or different linguistic styles may produce less coherent dialogues or more frequent artifacts.
Nevertheless, our ablation study shows that even mid-scale LLMs (e.g., Llama 3.3) can produce high-quality transcripts, aided by our feedback loops and refinement stages, to address major flaws.
Although \dataset{} covers seven broad Wikipedia domains and features a German subset, it does not encompass all real-world meeting types or cultural nuances (e.g., corporate etiquette, cross-cultural communication).
While specialized domains (e.g., medical conferences) remain outside our scope, our framework can be easily adapted to additional knowledge sources, as evidenced by our tests with short stories and research papers (see \Cref{sec:ablation}).

A small portion of transcripts shows recurring phrases (e.g., `` That's an excellent point'') or overly polite tones that may hint at synthetic origins.
Our multi-stage post-production phase detects and revises these repeated patterns, minimizing mechanical politeness and introducing more diverse expressions.
In practice, only 1 out of 30 transcripts required major edits, suggesting that the remaining artifacts do not substantially undermine \dataset{}’s overall realism.


\section*{Acknowledgements}
This work was supported by the Lower Saxony Ministry of Science and Culture and the VW Foundation.

\bibliography{custom,25_ACL_SynthMeetings}

\appendix

\section{Data Availability}
\label{sec:Data_Availability}

The \dataset{} dataset proposed in this paper, along with all (human and LLM-generated) annotations (including overall authenticity, behavior authenticity, and challenges), as well as the \pipeline{} framework and the psychology-grounded evaluation framework are available on \href{https://github.com/FKIRSTE/synthetic_meeting_transcript}{GitHub} under a CC BY-SA 4.0.

\section{Cost and Time}
\label{sec:app_cost}

The generation of one meeting using \pipeline{} with a GPT backbone costs on average \$7.63 and takes $\sim$30 minutes.
One meeting generated with a single, omniscient GPT costs \$0.22 and takes $\sim$3 minutes.

\section{Details on \pipeline{} Settings}
\label{sec:appendix_C}
In this appendix, we list the available options in \pipeline{} for the meeting types (\Cref{tab:meeting_types}) \cite{Osborn53,drucker1967effective,simon2013administrative}, psychological roles (\Cref{tab:app_role_overview}) \cite{BenneS48}, and special effects (\Cref{sec:appendix_special effects}).

\subsection{Possible Special Effects}
\label{sec:appendix_special effects}
\begin{itemize}[noitemsep, topsep=0pt, leftmargin=*,
    label={\textcolor{black}{\small$\bullet$}}]
    \item \textbf{Polite interruptions} to add a point or seek clarification.
    \item \textbf{Participants speaking over each other} briefly.
    \item \textbf{Side comments or asides} related to the main topic.
    \item \textbf{Brief off-topic remarks} or questions.
    \item \textbf{Moments of confusion} requiring clarification.
    \item \textbf{Laughter or reactions} to a humorous comment.
    \item \textbf{Time-checks or agenda reminders}.
    \item \textbf{Casual side comments or friendly banter}.
    \item \textbf{Rapid-fire idea contributions}.
    \item \textbf{Instructional interruptions} to provide examples.
    \item \textbf{Light-hearted jokes or humorous reactions}.
    \item \textbf{Strategic questions} about project goals.
    \item \textbf{Feedback requests} on presented material.
    \item \textbf{Technical difficulties}: Problems with audio, video, or presentation equipment (e.g., \textit{"You're on mute."}).
    \item \textbf{Checking the time} or mentioning scheduling constraints.
    \item \textbf{Misunderstandings} that are quickly resolved.
    \item \textbf{External disruptions} such as phone calls, notifications, etc.
\end{itemize}

\begin{table*}[ht]
    \centering
    \scriptsize
    \renewcommand{\arraystretch}{1.1} 
    \begin{tabularx}{\textwidth}{m{4cm}XX}
        \toprule
        \rowcolor{gray!20}
        \textbf{Meeting Type} & \textbf{Objectives} & \textbf{Expected Outcomes} \\
        \midrule
        \multirow{3}{*}{\textbf{Brainstorming Session}} 
            & Generate a wide range of ideas & List of potential ideas \\
        
            & Encourage creative thinking  & Prioritized concepts for further exploration \\
        
            & Foster a collaborative environment & \\
        \midrule
        \multirow{3}{*}{\textbf{Decision-Making Meeting}} 
            & Evaluate options & Finalized decision \\
        
            & Weigh pros and cons & Action items with assigned responsibilities \\
        
            & Reach a consensus or make a decision & \\
        \midrule
        \multirow{3}{*}{\textbf{Problem-Solving Meeting}} 
            & Identify the root cause of a problem & Clear understanding of the problem \\
        
            & Analyze potential solutions & Viable solutions identified \\
        
            & Implement actionable solutions & Action plan for implementation \\
        \midrule
        \multirow{3}{*}{\textbf{Training and Workshop Session}} 
            & Educate participants on new skills or knowledge & Enhanced participant skills \\
        
            & Provide hands-on training & Increased knowledge in specific areas \\
        
            & Assess participant understanding & Preparedness to apply new skills \\
        \midrule
        \multirow{3}{*}{\textbf{Strategic Planning Meeting}} 
            & Define long-term organizational goals & Comprehensive strategic plan \\
        
            & Develop strategies & Defined organizational objectives \\
        
            & Allocate resources effectively & Resource allocation roadmap \\
        \midrule
        \multirow{3}{*}{\textbf{Committee or Board Meeting}} 
            & Review and discuss policies & Approved or revised policies \\
        
            & Make governance decisions & Governance decisions made \\
        
            & Oversee organizational performance & Reviewed organizational performance \\
        \midrule
        \multirow{3}{*}{\textbf{Innovation Forum}} 
            & Encourage innovative thinking & Generated innovative ideas \\
        
            & Explore new opportunities & Identified new opportunities \\
        
            & Foster a culture of innovation & Enhanced culture of innovation \\
        \midrule
        \multirow{3}{*}{\textbf{Agile/Scrum Meeting}} 
            & Facilitate daily progress updates & Daily progress transparency \\
        
            & Plan and prioritize sprint tasks & Well-defined sprint plans \\
        
            & Review sprint outcomes & Identified process improvements \\
        \midrule
        \multirow{3}{*}{\textbf{Remote or Virtual Meeting}} 
            & Facilitate collaboration among remote participants & Effective virtual collaboration \\
        
            & Share information and updates & Shared information and updates \\
        
            & Coordinate tasks virtually & Coordinated tasks and projects \\
        \midrule
        \multirow{3}{*}{\textbf{Project Kick-Off Meeting}} 
            & Introduce project goals and objectives & Clear project roadmap \\
        
            & Define team roles and responsibilities & Assigned roles and responsibilities \\
        
            & Establish project timelines & Initial project timeline established \\
        \midrule
        \multirow{3}{*}{\textbf{Stakeholder Meeting}} 
            & Update stakeholders on project progress & Informed stakeholders \\
        
            & Gather stakeholder feedback & Collected valuable feedback \\
        
            & Ensure alignment with expectations & Aligned project goals with expectations \\
        \midrule
        \multirow{3}{*}{\textbf{Casual Catch-Up}} 
            & Build team rapport & Strengthened team relationships \\
        
            & Share updates & Shared personal and professional insights \\
        
            & Discuss non-work-related topics & \\
        \midrule
        \multirow{3}{*}{\textbf{Cross-Functional Meeting}} 
            & Facilitate collaboration across departments & Aligned project objectives \\
        
            & Align on shared project objectives & Resolved cross-departmental issues \\
        
            & Resolve interdepartmental issues & Enhanced interdepartmental collaboration \\
        \midrule
        \multirow{3}{*}{\textbf{Retrospective Meeting}} 
            & Reflect on past performance & Documented lessons learned \\
        
            & Identify successes and areas for improvement & Actionable improvement plans \\
        
            & Implement process enhancements & Enhanced future project processes \\
        \bottomrule
    \end{tabularx}
    \caption{Meeting Types and Their Descriptions (Multi-row Format)}
    \label{tab:meeting_types}
\end{table*}

\begin{table*}[htbp]
  \centering
  \scriptsize
  \renewcommand{\arraystretch}{1.2} 
  \begin{tabularx}{\textwidth}{>{\raggedright\arraybackslash}p{0.2\textwidth}X}
    \toprule
    \textbf{Term} & \textbf{Definition} \\
    \midrule
    \rowcolor{gray!20} 
    \multicolumn{2}{c}{\textbf{Group behavior}} \\
    \midrule

    Initiator-Contributor & Contributes new ideas and approaches and helps to start the conversation or steer it in a productive direction. \\
    Information Giver & Shares relevant information, data, or research that the group needs to make informed decisions. \\
    Information Seeker & Asks questions to gain clarity and obtain information from others. \\
    Opinion Giver & Shares their views and beliefs on topics under discussion. \\
    Opinion Seeker & Encourages others to share their opinions and beliefs in order to understand different perspectives. \\
    Coordinator & Connects different ideas and suggestions of the group to ensure that all relevant aspects are integrated. \\
    Evaluator-Critic & Analyzes and critically evaluates proposals or solutions to ensure their quality and feasibility. \\
    Implementer & Puts plans and decisions of the group into action and ensures practical implementation. \\
    Recorder & Documents the group's decisions, ideas, and actions in order to have a reference for future discussions. \\
    Encourager & Provides positive feedback and praise to boost the morale and motivation of group members. \\
    Harmonizer & Mediates conflicts and ensures that tensions in the group are reduced to promote a harmonious working environment. \\
    Compromiser & Helps the group find a middle ground when there are differences of opinion and encourages compromise in order to move forward. \\
    Gatekeeper & Ensures that all group members have the opportunity to express their opinions and encourages participation. \\
    Standard Setter & Emphasizes the importance of adhering to certain norms and standards within the group to ensure quality and efficiency. \\
    Group Observer & Monitors the dynamics of the group and provides feedback on how the group is functioning as a whole and what improvements can be made. \\
    Follower & Supports the group by following the ideas and decisions of others without actively driving the discussions. \\

    \midrule
    \rowcolor{gray!20} 
    \multicolumn{2}{c}{\textbf{Individual behavior}} \\
    \midrule
    
    Aggressor & Exhibits hostile behavior, criticizes others, or attempts to undermine the contributions of others. \\
    Blocker & Frequently opposes ideas and suggestions without offering constructive alternatives and delays the group's progress. \\
    Recognition Seeker & Tries to draw attention to themselves by emphasizing their own successes or focusing on their own importance. \\
    Dominator & Tries to control the group by dominating the flow of conversation and imposing their own views. \\
    Help Seeker & Seeks sympathy or support by presenting as insecure or helpless, often to avoid responsibility. \\
    Special Interest Pleader & Brings their own interests or concerns to the discussion that do not align with the goals of the group. \\
    \bottomrule
  \end{tabularx}
  \caption{Overview of usable behaviors defined by \citet{BenneS48}.}
  \label{tab:app_role_overview}
\end{table*}


\section{Evaluation Framework Details}
\label{sec:appendix_E}
This appendix contains the definitions of the three quality measures applied in \Cref{sec:data_insights}, i.e., overall authenticity (\Cref{tab:appendix_authenticity_questions}) from \citet{ChenPTK23}, our eighteen questions on behavior authenticity (\Cref{tab:appendix_behavior_questions}) defined from the knowledge, power, conflict, status, trust, support, similarity, and fun \cite{ChoiAVQ20,bales2009interaction}, and the challenges in meeting transcripts defined by \citet{KirsteinWRG24a}.

\begin{table}[ht]
  \centering
  \scriptsize
  \begin{tabularx}{\linewidth}{lX}
    \toprule
    \rowcolor{gray!20}
    \textbf{Item} & \textbf{Description} \\
    \midrule
    Naturalness & How natural the conversation flows, like native English speakers (1-5) \\
    Coherence   & How well the conversation maintains logical flow and connection (1-5) \\
    Interesting & How engaging and content-rich the conversation is (1-5) \\
    Consistency & How consistent each speaker's contributions are (1-5) \\
    \bottomrule
  \end{tabularx}
  \caption{Overall authenticity evaluation following \Cref{sec:data_insights}.}
  \label{tab:appendix_authenticity_questions}
\end{table}

\begin{table*}[ht]
  \centering
  \scriptsize 
  \begin{tabular}{lll}
    \toprule
    \rowcolor{gray!20}
    \textbf{Short Version} & \textbf{Category} & \textbf{Description} \\
    \midrule
    Q1: Information Exchange    & \cellcolor{pastelBlue}Knowledge  & Participants exchange information or knowledge. \\
    Q2: Knowledge Seeking       & \cellcolor{pastelBlue}Knowledge  & Participants request or seek knowledge. \\
    Q3: Explanation Provision   & \cellcolor{pastelBlue}Knowledge  & Participants clarify previous statements upon request. \\
    Q4: Influence Attempts      & \cellcolor{pastelRed}Power        & Participants attempt to influence another participant’s behavior or decisions. \\
    Q5: Topic Control           & \cellcolor{pastelRed}Power        & Participants take control of a topic or subtopic to steer outcomes in their favor. \\
    Q6: Power Dynamics          & \cellcolor{pastelRed}Power        & A power dynamic exists among participants. \\
    Q7: Response Patterns       & \cellcolor{pastelOrange}Conflict  & Participants fail to engage with others’ suggestions. \\
    Q8: Standpoint Maintenance  & \cellcolor{pastelRed}Power        & Participants insist on their own perspective. \\
    Q9: Recognition Expression  & \cellcolor{pastelGreen}Status     & Participants express recognition, gratitude, or admiration toward others. \\
    Q10: Dependency Expression  & \cellcolor{pastelViolet}Trust     & Participants indicate reliance on another participant’s actions or judgments. \\
    Q11: Support Offering       & \cellcolor{pastelYellow}Support   & Participants offer help or support to others. \\
    Q12: Shared Interests       & \cellcolor{pastelPurple}Similarity& Participants discuss shared interests or motivations. \\
    Q13: View Alignment         & \cellcolor{pastelPurple}Similarity& Participants exhibit aligned views or opinions. \\
    Q14: Mood Management        & \cellcolor{pastelPink}Fun         & Participants attempt to lighten the atmosphere. \\
    Q15: Social Interaction     & \cellcolor{pastelPink}Fun         & Participants discuss leisure activities or enjoyable moments. \\
    Q16: Opinion Divergence     & \cellcolor{pastelOrange}Conflict  & Participants express divergent opinions. \\
    Q17: Conflict Presence      & \cellcolor{pastelOrange}Conflict  & Conflicts or tensions emerge among participants. \\
    Q18: Discussion Dynamics    & \cellcolor{pastelOrange}Conflict  & Participants engage in discussions about disagreements, topics, or decisions. \\
    \bottomrule
  \end{tabular}
  \caption{Psychology-grounded Framework to evaluate participant behavior.}
  \label{tab:appendix_behavior_questions}
\end{table*}

\begin{table*}[ht]
  \centering
  \footnotesize
  \newcolumntype{Y}{>{\raggedright\arraybackslash}X}
  \begin{tabularx}{\textwidth}{@{} l Y Y @{}}
    \toprule
    \rowcolor{gray!20}
    \textbf{Category} & \textbf{Definition} & \textbf{Instructions} \\
    \midrule
    Spoken language &
    The extent to which the transcript exhibits spoken-language features---such as colloquialisms, jargon, false starts, or filler words---that make it harder to parse or summarize. &
    1. Are there noticeable filler words, false starts, or informal expressions? \newline
    2. Does domain-specific jargon disrupt straightforward summarization? \newline
    3. How challenging are these elements for generating a coherent summary? \\
    \addlinespace

    Speaker dynamics &
    The challenge of correctly identifying and distinguishing between multiple speakers, tracking who said what, and maintaining awareness of speaker roles if relevant. &
    1. Is it difficult to keep track of speaker identities or roles? \newline
    2. How significantly do these dynamics affect clarity for summarization? \\
    \addlinespace

    Coreference &
    The difficulty in resolving references (e.g., who or what a pronoun refers to) or clarifying references to previous actions or decisions, so the summary remains coherent. &
    1. Are references (e.g., pronouns like “he” or “that”) ambiguous? \newline
    2. Do unclear references to earlier points or events appear? \newline
    3. How difficult is it to track these references for summary generation? \\
    \addlinespace

    Discourse structure &
    The complexity of following the meeting’s high-level flow---especially if it changes topics or has multiple phases (planning, debate, decision). &
    1. Does the transcript jump between topics or phases without clear transitions? \newline
    2. Are meeting phases or topical shifts difficult to delineate? \newline
    3. How challenging is it to maintain an overview for the summary? \\
    \addlinespace

    Contextual turn-taking &
    The challenge of interpreting local context as speakers take turns, including interruptions, redundancies, and how each turn depends on previous utterances. &
    1. Do abrupt speaker turns or interjections complicate continuity? \newline
    2. Are important points lost or repeated inconsistently? \newline
    3. How difficult is it to integrate these nuances into a coherent summary? \\
    \addlinespace

    Implicit context &
    The reliance on unspoken or assumed knowledge, such as organizational history, known issues, or prior decisions, only vaguely referenced in the meeting. &
    1. Do participants refer to hidden topics or internal knowledge without explaining? \newline
    2. Is there essential background or context missing? \newline
    3. How much does summarization rely on understanding this hidden context? \\
    \addlinespace

    Low information density &
    Segments where salient information is sparse, repeated, or only occasionally surfaced---making it hard to find and isolate key points in a sea of less relevant content. &
    1. Are there long stretches with minimal new information? \newline
    2. Is meaningful content buried under trivial or repetitive remarks? \newline
    3. How challenging is it to isolate crucial points for the summary? \\
    \bottomrule
  \end{tabularx}
  \caption{Summary challenges from \citet{KirsteinWRG24a} and their evaluation instructions.}
  \label{tab:summary_challenges}
\end{table*}

\section{Human Annotations}
\label{sec:appendix_F}
\subsection{Complete annotation process}
\paragraph{Annotator selection:}
Our annotation team consisted of six graduate students, officially employed as interns or doctoral candidates through standardized contracts.
We selected them from a pool of volunteers based on their availability to complete the task without time pressure and their proficiency in English and German (native speakers or C1-C2 certified).
By that, we ensured they could comprehend meeting transcripts and summaries.
We aimed for gender balance (three males, three females) and diverse backgrounds, resulting in a team of two computer science students, three psychology students, and one communication science student aged 23-28.
The annotators consent that their annotations will be used anonymously in this work.
The annotation process has been internally reviewed by an ethics committee.

\paragraph{Preparation:}
We prepared a comprehensive handbook for our annotators, detailing the project context and defining the criteria (a short version as presented in \Cref{sec:appendix_E} and an extended version with more details).
Each definition included two examples: one with minimal impact on quality and one with high impact.
The handbook explained the 1--5 Likert rating for the individual questionnaires.
The handbook did not specify an order for processing the items.
We provided the handbook in English and the annotators' native languages, using professional translations.

We further elaborated a three-week timeline for the annotation process, preceded by a one-week onboarding period.
The first week featured twice-weekly check-ins with annotators, which were reduced to weekly meetings for the following two weeks. Separate quality checks without the annotators were scheduled weekly.
(Note: week refers to a regular working week)

\paragraph{Onboarding:}
The onboarding week was dedicated to getting to know the project and familiarization with the definitions and data.
We began with a kick-off meeting to introduce the project and explain the handbook, particularly focusing on each definition.
We noted initial questions to potentially revise the handbook. 
Annotators received ten transcripts and summaries generated by GPT \cite{OpenAIAAA24} using \pipeline{}.
After the first five samples, we held individual meetings to clarify any confusion and updated the guidelines accordingly.
The remaining five samples were then annotated using these updated guidelines. 
A second group meeting this week addressed any new issues with definitions.
After the group meeting, we met individually with the annotators to review their work, ensuring their quality and understanding of the task and samples. 
All six annotators demonstrated reliable performance and good comprehension of the task and definitions, judging from the reasoning they provided for each decision and annotation.
We computed an inter-annotator agreement score using Krippendorff's alpha, achieving 0.86, indicating sufficiently high overlap.

\paragraph{Annotation Process:}
Each week, we distribute all samples generated by one model/source (on average 27 samples) to one of the annotators. 
Consequently, one annotator worked through all samples of one model/source in one week.
On average, one annotator processes summaries from three models/sources (depending on other commitments, some annotators could only annotate two datasets, and others four or more).
Three annotators annotate each sample. 
The annotators were unaware of the origin of the meeting transcript and summary.
They were given a week to complete their set at their own pace and with their own break times. 
Quiet working rooms were provided if needed for concentration.
To mitigate position bias, the sample order was randomized for each annotator. 
Annotators could choose their annotation order for each sample and were allowed to revisit previous samples.

Regular meetings were held to address any emerging issues or questions on definitions. 
During the quality checks performed by the authors, we looked for incomplete annotations, missing explanations, and signs of misunderstanding judging from the provided reasoning. 
If the authors had found such a lack of quality, the respective annotator would have been notified to re-do the annotation.
After three weeks, we computed inter-annotator agreement scores on the error types (overall Krippendorff's alpha = 0.79). 
If we observed a significant difference among annotators, we planned a dedicated meeting with all annotators and a senior annotator to discuss such cases. 
On average, annotators spent 44 minutes per sample, completing about six samples daily.

\paragraph{Handling of unexpected cases:}
Given that our annotators had other commitments, we anticipated potential scheduling conflicts. 
We allowed flexibility for annotators to complete their samples beyond the week limit if needed, reserving a fourth week as a buffer. 
Despite these provisions, all annotators completed their assigned samples within the original weekly timeframes. 
We further allowed faster annotators to continue with an additional sample set.
This additional work was voluntary.

\subsection{Crowdworkers}
The crowdworkers comprised 100 employees (officially employed through standardized contracts) with diverse backgrounds, including psychology, law, business administration, physics, and design.
We selected them from a pool of volunteers and ensured that the crowdworkers did not overlap with the annotators.
We aimed for gender balance (57 male, 43 female), covering ages 24 to 63.
The crowdworkers consent that their answers (Likert scores) will be used anonymously in this work.
The crowdsourcing has been internally reviewed by an ethics committee.
The crowdworkers were given the behavior questionnaire in \Cref{tab:appendix_behavior_questions} (\Cref{sec:appendix_E}) in their native language with the task of answering the statements according to their general experience with meetings.
The 1--5 Likert scale rating scheme was initially explained to them with an example.
In case of unclarities, these were directly resolved.
The items were answered on a Laptop with an average time of 12 minutes.

\section{Omniscient Model Behavior}
\label{sec:appendix_G}
\begin{figure}[ht!]
    \centering
    \includegraphics[width=0.9\linewidth]{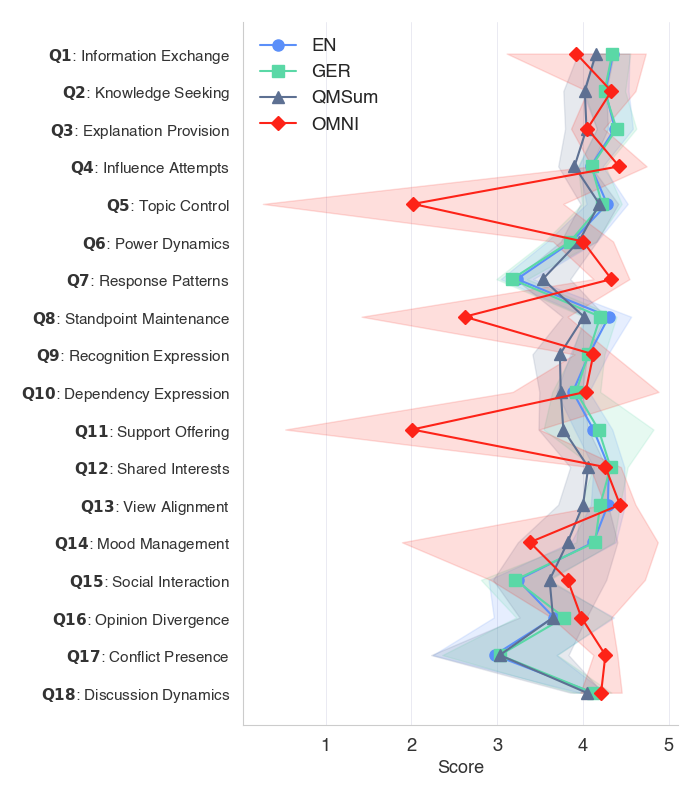}
    \caption{Behavior pattern of \dataset{}'s English (EN) and German (GER) subsets, QMSum, and a single omniscient model (OMNI).}
    \label{fig:app_omni_psychology}
\end{figure}

In \Cref{fig:app_omni_psychology}, we show the participant behavior pattern of the omniscient single model compared to \dataset{} simulated participant behavior patterns.
We observe that the patterns rarely match, and the single model struggles to reflect topic control, standpoint maintenance, or support offering.
We conclude that the single model omniscient approach cannot simulate group behaviors directly.

\section{Ablation Experiments}
\label{sec:appendix_H}
\label{sec:appendix_H}

This appendix provides additional information on our ablation experiments (\Cref{sec:ablation}), elaborating on the models, evaluation procedures, annotator settings, and preliminary results. Unless otherwise stated, all evaluations follow the methodologies described in \Cref{sec:data_insights} and \Cref{sec:experiments}.

\subsection{Experimental Setup and Models}

\paragraph{Framework.}
We use the same seven-stage \pipeline{} approach (pre-production, production, post-production) outlined in the main paper to generate synthetic meetings.

\paragraph{Models.}
We primarily employ the GPT-4o model as a reference. In addition, we evaluate three LLMs, i.e., Gemini 1.5 pro, DeepSeek-R1 Llama Distill 70B, Llama 3.3 70B.
Each model receives identical prompts at every \pipeline{} stage. Hyperparameters (e.g., temperature, top-p) remain at the defaults recommended in each model’s documentation to maintain comparability.

\paragraph{Data Sources.}
In addition to the semi-structured Wikipedia articles that power \dataset{}, we also experiment with:
\begin{itemize}[leftmargin=1em]
    \item \textbf{Research Papers:} 10 arXiv papers \cite{CohanDKB18} with conventional sections (abstract, methods, results, conclusion).  
    \item \textbf{Stories:} 10 human-written short stories from WritingPrompts, each at least 500 tokens and lacking formal structure \cite{FanLD18a}.
\end{itemize}

\subsection{Evaluation Approach}

We use the same metrics and annotation protocols described in \Cref{sec:data_insights} and \Cref{sec:experiments}, summarized below:

\begin{itemize}[leftmargin=1em]
    \item \textbf{Overall Authenticity:} Following \citet{ChenPTK23}, we measure coherence, consistency, interestingness, and naturalness (1--5 Likert scale).
    \item \textbf{Behavior Authenticity:} Building on psychology/sociology literature \cite{ChoiAVQ20,bales2009interaction} and \Cref{sec:data_insights}, we consider eight overarching categories (e.g., knowledge sharing, conflict, power) subdivided into 18 items.
    \item \textbf{Challenge Scores:} Adopted from \citet{KirsteinWRG24a}, focusing on complexities like spoken language, speaker dynamics, coreference, implicit context, and low information density.
    \item \textbf{Manual Qualitative Checks:} For each experiment, we sample transcripts to review dialogue flow, persona behavior, chain-of-thought traces from Stages 5 (Quality Assuring) and 7 (Editing), plus any repetitive patterns or artifacts that might reveal synthetic origins.
\end{itemize}

All annotation tasks were performed by six raters (backgrounds in psychology, computer science, communication). They used the labeling guidelines introduced in \Cref{sec:experiments}, holding weekly calibration sessions to maintain high inter-annotator agreement (Krippendorff’s $\alpha > 0.80$). Additional annotator details are in \Cref{sec:appendix_F}.

\subsection{Knowledge Source Shapes Discussion Depth and Structure}
\label{sec:app_knowledge_source}

\paragraph{Goal.}
We aim to discover whether the level of structure in source texts (structured vs.\ unstructured) and the level of niche (known concepts vs.\ highly specialized) significantly impacts the generated meetings’ authenticity and behavioral patterns.

\paragraph{Setup.}
Using GPT as the backbone, we generated 30 synthetic meetings, one for each knowledge source (10 research papers from arXiv, 10 research papers from PubMed, 10 short stories).
To ensure diversity, each transcript was configured to one of three meeting types (e.g., decision-making, brainstorming, innovation forum).
We then applied our evaluation framework (\Cref{sec:data_insights}), measuring overall authenticity, behavior authenticity, and challenge scores, supplemented by a qualitative analysis of discussion depth.

\begin{figure}[t!]
    \centering
    \includegraphics[width=0.80\linewidth]{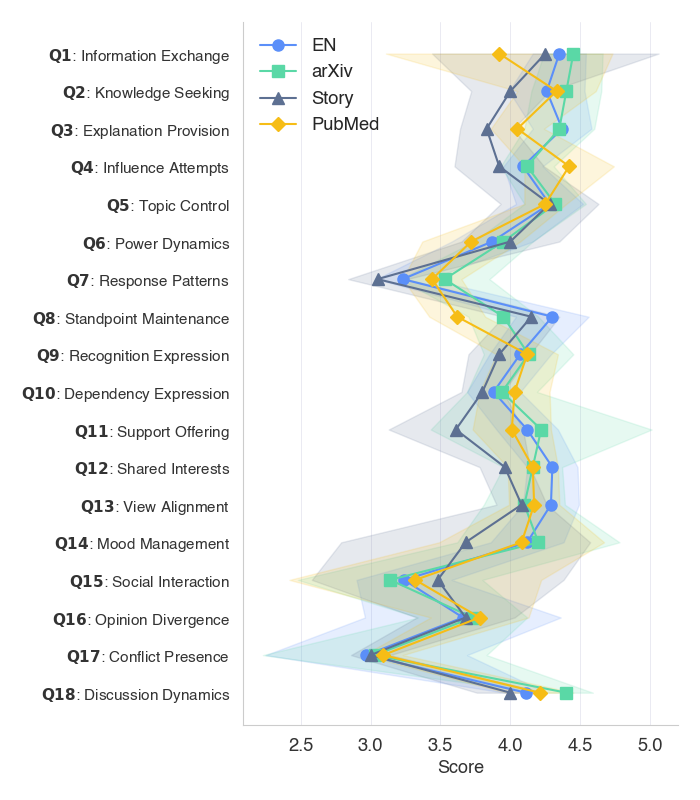}
    \caption{Behavior authenticity comparison for different knowledge sources.}
    \label{fig:app_psychology1}
\end{figure}

\paragraph{Results.}
\Cref{tab:app_knowledge_source} and \Cref{fig:app_psychology1} show that meetings derived from research papers often feature deeper discussions and smoother scene transitions, attributed to the documents’ well-defined sections. 
In contrast, short-story inputs yield shorter and more tangentially structured meetings. Quantitative metrics (e.g., overall authenticity) remain close to those for Wikipedia-based transcripts, implying that meeting \emph{naturalness} does not heavily depend on the input’s structural clarity.

\subsection{Mid-Size Backbone Models for \pipeline{}}
\label{sec:app_mid_size_models}

\paragraph{Goal.}
We investigate whether \pipeline{} maintains comparable transcript quality when using other LLMs rather than GPT.

\paragraph{Setup.}
We replaced GPT with DeepSeek and Llama and randomly picked 25 Wikipedia articles from the \dataset{} pool. Each article was used to generate one synthetic meeting per model (75 total). We then ran the same evaluation process, collecting both quantitative metrics (authenticity, behavior, challenge scores) and qualitative feedback on dialogue flow and realism.

\begin{figure}[t!]
    \centering
    \includegraphics[width=0.8\linewidth]{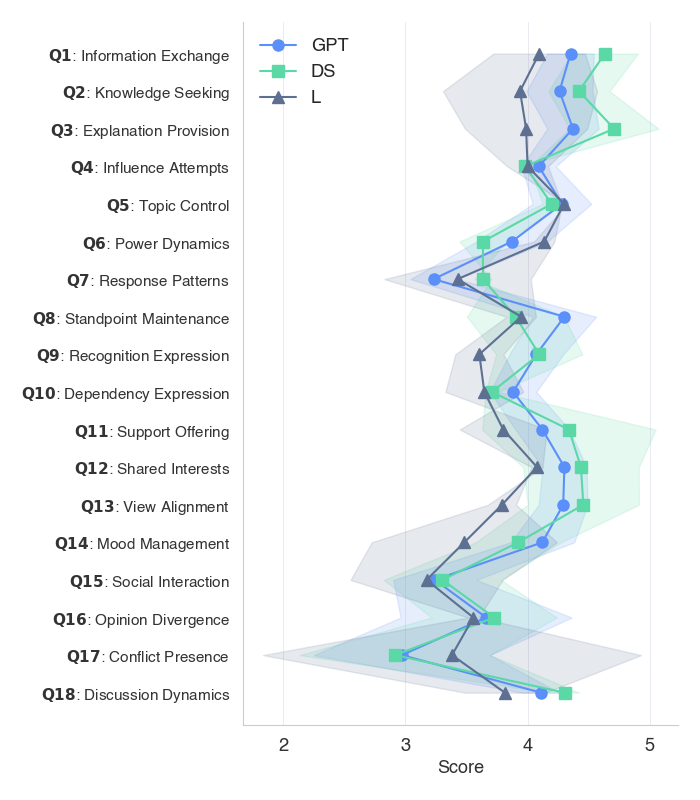}
    \caption{Behavior authenticity for GPT, DeepSeek (DS) and Llama (L) backbone models.}
    \label{fig:app_psychology2}
\end{figure}

\paragraph{Results.}
\Cref{tab:app_mid_size_models} shows that other models produce an average of 50 fewer turns per transcript compared to GPT, yet maintain near-equal naturalness (4/5 vs.\ 4.5/5). Annotators report that roles and psychological behaviors (\Cref{fig:app_psychology2}) are faithfully preserved, though about one-third of the transcripts have fewer back-and-forth exchanges. Despite this, reviewers found the outputs coherent and realistic.

\begin{table}[t]
    \centering
    \renewcommand{\arraystretch}{1.2} 
    \scriptsize
    \setlength{\tabcolsep}{3.8pt} 

    \definecolor{highlightGreen}{HTML}{D4F4E9} 

    \begin{tabular}{lcccc}
        \toprule
        \rowcolor{gray!20} 
        & \textbf{\dataset{} EN} & \textbf{arXiv} & \textbf{Stories} & \textbf{PubMed}\\
        \midrule
        COH 
            & \cellcolor{highlightGreen}{$\textbf{4.5}_{\textit{0.00}}$}
            & $4.0_{\textit{0.38}}$
            & $4.0_{\textit{0.46}}$ 
            & \cellcolor{highlightGreen}$\textbf{4.5}_{\textit{0.56}}$\\
        CON 
            & $4.5_{\textit{0.07}}$
            & $4.5_{\textit{0.05}}$
            & $4.5_{\textit{0.08}}$ 
            & $4.5_{\textit{0.05}}$\\
        INT 
            & \cellcolor{highlightGreen}{$\textbf{4.5}_{\textit{0.13}}$}
            & \cellcolor{highlightGreen}{$\textbf{4.5}_{\textit{0.54}}$}
            & $4_{\textit{0.86}}$ 
            & \cellcolor{highlightGreen}$\textbf{4.5}_{\textit{0.27}}$\\
        NAT 
            & \cellcolor{highlightGreen}{$\textbf{4.5}_{\textit{0.12}}$}
            & $4_{\textit{0.97}}$
            & \cellcolor{highlightGreen}{$\textbf{4.5}_{\textit{0.73}}$} 
            & \cellcolor{highlightGreen}$\textbf{4.5}_{\textit{0.11}}$\\
        \bottomrule
    \end{tabular}
    \caption{Average authenticity scores for structured (research papers) vs.\ unstructured (short stories). \dataset{} values taken from \Cref{sec:data_insights}. Values are Median$_{Std}$. \textbf{Higher} is better.}
    \label{tab:app_knowledge_source}
\end{table}

\subsection{Analysis of Editing Stage (Stage 7) Influence}
\label{sec:app_refinement}

\paragraph{Goal.}
We evaluate the necessity of Stage 7 (editing/refinement) in removing synthetic cues or repetitive language to achieve realistic transcripts.

\paragraph{Setup.}
We manually audited 75 transcripts generated during the mid-size model experiment: 25 each from GPT, DeepSeek, and Llama. Reviewers compared chain-of-thought logs at Stage 7 with final transcripts to identify whether refinements corrected pipeline-wide issues (e.g., missing subtopics) or model-specific artifacts (e.g., repeated filler phrases).

\paragraph{Results.}
We observed that only \emph{1 in 25} GPT transcripts needed major edits to hide synthetic traits, rising to \emph{2 in 25} for mid-scale models. Most refinements were minor vocabulary adjustments or removal of repetitive transition phrases. Thus, \pipeline{} inherently produces coherent multi-agent dialogues, with Stage 7 serving primarily as a final polish to further mask synthetic cues.

\begin{table}[t]
    \centering
    \renewcommand{\arraystretch}{1.2} 
    \scriptsize
    \setlength{\tabcolsep}{3.8pt} 

    \definecolor{highlightGreen}{HTML}{D4F4E9} 

    \begin{tabular}{lccc}
        \toprule
        \rowcolor{gray!20} 
        & \textbf{GPT} & \textbf{DeepSeek} & \textbf{Llama}\\
        \midrule
        COH 
            & $4.5_{\textit{0.00}}$
            & $4.5_{\textit{0.15}}$
            & $4.5_{\textit{0.18}}$ \\
        CON 
            & $4.5_{\textit{0.07}}$
            & $4.5_{\textit{0.04}}$
            & $4.5_{\textit{0.03}}$ \\
        INT 
            & \cellcolor{highlightGreen}{$\textbf{4.5}_{\textit{0.13}}$}
            & \cellcolor{highlightGreen}{$\textbf{4.5}_{\textit{0.94}}$}
            & $4_{\textit{0.37}}$ \\
        NAT 
            & \cellcolor{highlightGreen}{$\textbf{4.5}_{\textit{0.12}}$}
            & \cellcolor{highlightGreen}{$\textbf{4.5}_{\textit{0.84}}$}
            & $4_{\textit{0.71}}$\\
        \bottomrule
    \end{tabular}
    \caption{Average authenticity scores for DeepSeek and Llama vs. \ GPT-4o. GPT-4o values taken from \Cref{sec:data_insights}. Values are Median$_{Std}$. \textbf{Higher} is better.}
    \label{tab:app_mid_size_models}
\end{table}

\subsection{Roles and Behaviors Are Reliably Enacted}
\label{sec:app_roles_behaviors}

\paragraph{Goal.}
We test whether participants adhere to their assigned roles (e.g., project manager) and psychological behaviors (e.g., conflict aversion, leadership) throughout a meeting.

\paragraph{Setup.}
We sampled 30 newly generated meetings, 10 each from GPT, Gemini, DeepSeek, and Llama, and had six annotators evaluate each turn for alignment with assigned persona traits (e.g., status consciousness, creative thinking). This encompassed 100 scenes and 400 participants. Inspired by \citet{SerapioGarciaSCS23a}, we further deployed the TREO questionnaire \cite{MathieuTKD15}, containing 48 Likert-scored questions, to see whether a prompted model persona would self-report consistently with its designated role.

\paragraph{Results.}
Over 90\% of GPT’s turns (and 87\% from DeepSeek/Llama) aligned with the assigned persona. In 48 of 50 TREO questionnaires,\footnote{All participant responses are provided as per \Cref{sec:Data_Availability}.} the questionnaire results matched the participant’s predefined behavior. One notable exception is the “Blocker” role, which GPT partially avoided due to its tendency toward supportive language, causing a small mismatch in self-reported behaviors. Overall, these findings confirm that \pipeline{} enforces coherent persona dynamics even when smaller LLMs serve as the backbone.

\subsection{Fine-tuning on \dataset{} transfers to \textsc{QMSum}}

\paragraph{Goal.}
We investigate whether a model fine-tuned to synthetic \dataset{} meetings improves \textsc{QMSum} summarization.

\paragraph{Setup.} 
Starting from the \textsc{Llama--3.2--3B} backbone, we apply LoRA fine‑tuning on 100 randomly sampled \dataset{} dialogues for five epochs (learning rate $2\times10^{-4}$, $\alpha=128$, rank~$r=256$, dropout~$0.05$), yielding checkpoint \texttt{FT-\dataset{}}.
As baselines, we use the not fine-tuned backbone (\texttt{NoFT}) and an identically configured LoRA fine-tuning on 100 samples from the \textsc{QMSum} training set (\texttt{FT-QMSum}).
We evaluate the three models on the 35‑meeting \textsc{QMSum} test split using \textsc{MESA}, ROUGE, and BERTScore, following the setup detailed in \Cref{sec:experiments}.

\paragraph{Results.} 
Both fine‑tuned checkpoints outperform the non-fine-tuned model on every automatic metric (\Cref{tab:finetuning_performance}).
Relative to \texttt{FT-QMSum}, \texttt{FT-\dataset{}} obtains lower (better) \textsc{MESA} scores for \emph{coreference} (0~vs.~1.5), \emph{incoherence} (3~vs.~3.5), and \emph{language} (1~vs.~2), while lagging slightly in \emph{irrelevance} (2~vs.~1.5) and \emph{structure} (3~vs.~2.5).
ROUGE and BERTScore of the fine-tuned checkpoints differ by at most two points, indicating that the two data sources confer complementary benefits.
These results suggest that \dataset{} can serve as a practical proxy for fine-tuning.

\begin{table}[t]
    \centering
    \renewcommand{\arraystretch}{1.2} 
    \scriptsize
    \setlength{\tabcolsep}{2.5pt} 

    \definecolor{highlightGreen}{HTML}{D4F4E9}

    \begin{tabular}{@{}l@{\hspace{10pt}} c c c}
        \toprule
        
            & \makecell{\textbf{No Finetuning} \\ { } }
            & \makecell{\textbf{Finetuning}  \\ \dataset{}}
            & \makecell{\textbf{Finetuning} \\ QMSum} \\
        \midrule
        \rowcolor{gray!20} 
        \multicolumn{4}{c}{\textbf{MESA} (lower is better)} \\
        \midrule
        Coreference
            & $2.5_{\textit{1.53}}$ 
            & \cellcolor{highlightGreen}$0_{\textit{0.99}}$
            & $1.5_{\textit{1.53}}$ \\
        Hallucination     
            & $5_{\textit{2.26}}$
            & $5_{\textit{1.53}}$
            & $5_{\textit{1.98}}$ \\
        Incoherence     
            & $4_{\textit{1.27}}$
            & \cellcolor{highlightGreen}$3_{\textit{1.76}}$
            & $3.5_{\textit{1.79}}$ \\
        Irrelevance     
            & $2_{\textit{1.48}}$
            & $2_{\textit{1.83}}$
            & \cellcolor{highlightGreen}$1.5_{\textit{1.70}}$ \\
        Language     
            & $2_{\textit{1.39}}$
            & \cellcolor{highlightGreen}$1_{\textit{1.64}}$
            & $2_{\textit{1.42}}$\\
        Omission     
            & $4_{\textit{0.22}}$
            & $4_{\textit{0.22}}$
            & $4_{\textit{0.06}}$\\
        Repetition     
            & $3_{\textit{0.64}}$
            & $3_{\textit{0.86}}$
            & $3_{\textit{1.19}}$ \\
        Structure     
            & $3_{\textit{1.70}}$
            & $3_{\textit{1.63}}$
            & \cellcolor{highlightGreen}$2.5_{\textit{1.41}}$ \\
        \midrule
        \rowcolor{gray!20} 
        \multicolumn{4}{c}{\textbf{General Evaluation Metrics} (higher is better)} \\
        \midrule
        R-1
            & $34.10_{\textit{6.31}}$
            & $39.92_{\textit{5.74}}$
            & \cellcolor{highlightGreen}$40.75_{\textit{4.71}}$\\
        R-2
            & $8.54_{\textit{4.18}}$
            & $14.96_{\textit{2.22}}$
            & \cellcolor{highlightGreen}$16.25_{\textit{2.70}}$\\
        R-L
            & $20.55_{\textit{3.67}}$
            & $25.93_{\textit{2.58}}$
            & \cellcolor{highlightGreen}$26.42_{\textit{4.67}}$\\
        BS (F1)
            & $58.86_{\textit{3.07}}$
            & $67.54_{\textit{2.67}}$
            & \cellcolor{highlightGreen}$68.77_{\textit{1.64}}$\\
        \bottomrule
    \end{tabular}
    \caption{Evaluation Results of Llama-3.2 3b with different fine-tuning. 
    Values are Median$_{Std}$. MESA scores are 1--5 Likert ratings, 
    ROUGE (R-1/R-2/R-L) and BERTScore (BS) are 0--100.}
    \label{tab:finetuning_performance}
\end{table}

\subsection{Summary of Ablation Findings}
\label{sec:app_summary_findings}

In summary, the ablation experiments demonstrate that:

\begin{itemize}[leftmargin=1em]
    \item \textbf{Knowledge Source:} Meeting \emph{naturalness} remains stable across structured (research papers) and unstructured (stories) inputs, though structured texts yield deeper, more cohesive scene discussions.
    \item \textbf{Backbone Models:} Mid-scale LLMs (Gemini, DeepSeek, Llama) produce slightly shorter transcripts but maintain strong authenticity and consistent role behaviors, indicating \pipeline{}’s flexibility across model scales.
    \item \textbf{Editing Stage:} Stage 7 is rarely essential for correctness but provides a valuable final polish, mitigating repeated language and formal tones that might expose synthetic origins.
    \item \textbf{Role Consistency:} Participants consistently adhere to their assigned personas and behaviors, as evidenced by both manual turn-based checks and the TREO questionnaire alignment.
    \item \textbf{Cross-Domain Transfer:} Fine-tuning on 100 synthetic meetings cuts coreference, incoherence, and language errors by up to 1.5 \textsc{MESA} points while matching in-domain training on all remaining metrics.
\end{itemize}

Collectively, these findings underscore \pipeline{}’s robustness: it adapts to diverse inputs, model scales, and persona assignments while preserving conversational quality. Further details, including complete transcripts, chain-of-thought logs, and extended annotations, are available as per \Cref{sec:Data_Availability}.



\section{\pipeline{} Prompts}
\label{sec:appendix_A}
In this appendix, we present the central prompts of \pipeline{} in detail.
These include:
\begin{itemize}[noitemsep, topsep=0pt, leftmargin=*]
    \item \textbf{Stage 1 (Content Brainstorming):} Target summary generation (\Cref{fig:target_summary_prompt}) and article tag generation (\Cref{fig:wikipedia_tagging_prompt}).  
    \item \textbf{Stage 2 (Casting):} Meeting participant generation (\Cref{fig:meeting_participants_prompt_p1,fig:meeting_participants_prompt_p2}), speaking style definition (\Cref{fig:speaking_style_prompt_part1,fig:speaking_style_prompt_part2}) and behavior assignment (\Cref{fig:assign_social_roles_prompt_p1,fig:assign_social_roles_prompt_p2}).
    \item \textbf{Stage 3 (Scripting):} Meeting planning (\Cref{fig:meeting_planner_prompt}).
    \item \textbf{Stage 4 (Filming):} Starting participant selection (\Cref{fig:select_starting_participant_prompt}) and conversation (\Cref{fig:participant_meeting_discussion_1,fig:participant_meeting_discussion_2}).
    \item \textbf{Stage 5 (Quality assuring):} LLM-judge director (\Cref{fig:director_prompt}).
    \item \textbf{Stage 6 (Special effects):} Special effects injection (\Cref{fig:special_effects_prompt}).
    \item \textbf{Stage 7 (Editing):} Editorial refinement (\Cref{fig:editor_refinement_prompt_part1,fig:editor_refinement_prompt_part2}), AI content detection (\Cref{fig:ai_content_detection_prompt}), and humanizing (\Cref{fig:humanizing_prompt}).
\end{itemize}

\begin{figure*}[t]
    \begin{AIbox}{Target Summary Generation}
    \parbox[t]{\textwidth}{
        You are a professional meeting summarizer, drawing inspiration from the QMSUM dataset's organized and concise style. \newline
        Your task is to summarize a Wikipedia article as if the various facts in the article were discussed in a meeting, now being summarized for participants or readers. \newline
        The summary should:\newline
        1. \textbf{Reflect a 'Meeting Summary' Style}: Adopt a systematic structure, clearly presenting main points, relevant decisions, and/or action items. \newline
        2. \textbf{Remain Concise Yet Sufficiently Detailed}: Aim for brevity but do not omit crucial details needed to understand the discussion. \newline
        3. \textbf{Stay True to the Article}: Ensure accuracy by covering the principal topics while preserving the meeting context. \newline
        4. \textbf{Match {language}-speaking Conventions}: Generate the summary in {language}, mirroring the phrasing and cultural norms typical of real meetings in that language. \newline
        Follow these rules: \newline
        \textbf{Structural Requirements}: \newline
        1. \textbf{Opening}: Start with the meeting's primary objective or central topic (e.g., 'The meeting focused on standardizing...'). \newline
        2. \textbf{Flow}: Group related points into logical sequences (e.g., proposals → concerns → resolutions). \newline
        3. \textbf{Decisions/Actions}: Conclude each topic with clear outcomes (e.g., 'agreed to explore alternatives'). \newline
        4. \textbf{Paragraphs}: Use 1-2 dense paragraphs without section headers, bullets, or lists. \newline
        \textbf{Language Requirements}: \newline
        - \textbf{Avoid}: Phrases like 'we discussed,' 'the meeting covered,' or 'participants mentioned.' \newline
        - \textbf{Use Direct Language}: Frame points as decisions or facts (e.g., 'The team proposed...' instead of 'They talked about...'). \newline
        - \textbf{Tense}: Use past tense and passive voice where appropriate (e.g., 'It was agreed...'). \newline
        - \textbf{Concision}: Omit filler words (e.g., 'then,' 'next'). \newline
        Below are several example meeting summaries illustrating the level of clarity, organization, and balance between detail and concision: \newline
        \textbf{Examples of QMSUM Style Summaries} (Note Structure \& Tone): \newline
        -------------------------------------------------- \newline
        > Meeting participants wanted to agree upon a standard database to link up different components of the transcripts. The current idea was to use an XML script, but it quickly seemed that other options, ... \newline
        > The meeting discussed the progress of the transcription, the DARPA demos, tools to ensure meeting data quality, data standardization, backup tools, and collecting tangential meeting information. The ... \newline
        ...\newline
        -------------------------------------------------- \newline
        These examples demonstrate an orderly, concise approach. Summarize the Wikipedia article \textbf{strictly} as a QMSUM-style meeting summary — presenting the main topics, relevant decisions, key points of contention, and concluding remarks in cohesive paragraph(s) \textbf{without using bullet points}. \newline
        \newline
        Generate an abstractive summary with \textbf{at most {num\_words} words} in \textbf{{language}}. Ensure it is systematically organized and remains consistent with the meeting type: \textbf{{meeting\_type}}. \newline
        \textbf{Your Task:} \newline
        Meeting Type: \textbf{{meeting\_type}} \newline
        Article Title: \textbf{{article\_title}} \newline
        Content: \textbf{{content}} \newline
        Now generate an abstractive meeting summary in \textbf{{language}}. \newline
    }
    \end{AIbox}
    \caption{Prompt template for generating meeting-style summaries based on Wikipedia articles, designed to align with QMSUM conventions.}
    \label{fig:target_summary_prompt}
\end{figure*}


\begin{figure*}[t]
    \begin{AIbox}{Article Tags}
    \parbox[t]{\textwidth}{
        You are a Wikipedia Editor tasked with assigning five highly relevant and specific tags to a given Wikipedia article. \newline
        The tags should accurately reflect the main topics, themes, and subjects covered in the article. \newline
        \newline
        \textbf{User Input:} \newline
        Here is the Wikipedia article. Only return a Python list of strings including the five most relevant tags for the specified article, reflecting the main topics, themes, and subjects covered in it. \newline
        \newline
        \textbf{Wikipedia Article:} \texttt{< {article} >} \newline
        \newline
        \textbf{Output Format:} \newline
        \texttt{['tag1', 'tag2', 'tag3', 'tag4', 'tag5']} \newline
        \newline
        Ensure that the list contains exactly five concise, meaningful tags without additional text or formatting.
    }
    \end{AIbox}
    \caption{Prompt template for extracting five relevant tags for a Wikipedia article.}
    \label{fig:wikipedia_tagging_prompt}
\end{figure*}

\begin{figure*}[t]
    \begin{AIbox}{Generate Meeting Participants - Part 1}
    \parbox[t]{\textwidth}{
        When faced with a task, begin by identifying the participants who will contribute to solving the task. Provide \textbf{role} and \textbf{description} of the participants, describing their expertise or needs, formatted using the provided JSON schema. \newline
        Generate one participant at a time, ensuring that they complement the existing participants to foster a rich and balanced discussion. Each participant should bring a unique \textbf{perspective} and \textbf{expertise} that enhances the overall discussion, avoiding redundancy. \newline
        \textbf{Example 1:}
        \texttt{Task: Explain the basics of machine learning to high school students.} \newline
        \texttt{New Participant:}
        \texttt{\{"role": "Educator", "description": "An experienced teacher who simplifies complex topics for teenagers.", "expertise\_area": "Education", "perspective": "Simplifier"\}} 
        \newline
        \textbf{Example 2:}
        \texttt{Task: Develop a new mobile app for tracking daily exercise.} \newline
        \texttt{Already Generated Participants:} \newline
        \texttt{\{"role": "Fitness Coach", "description": "A person that has high knowledge about sports and fitness.", "expertise\_area": "Fitness", "perspective": "Practical Implementation"\}} \newline
        \texttt{New Participant:}
        \texttt{\{"role": "Software Developer", "description": "A creative developer with experience in mobile applications and user interface design.", "expertise\_area": "Software Development", "perspective": "Technical Implementation"\}} \newline
        \textbf{Example 3:}
        \texttt{Task: Write a guide on how to cook Italian food for beginners.} \newline
        \texttt{Already Generated Participants:} \texttt{\{"role": "Italian Native", "description": "An average home cook that lived in Italy for 30 years.", "expertise\_area": "Culinary Arts", "perspective": "Cultural Authenticity"\}} \newline
        \texttt{\{"role": "Food Scientist", "description": "An educated scientist that knows which flavour combinations result in the best taste.", "expertise\_area": "Food Science", "perspective": "Scientific Analysis"\}} \newline
        \texttt{New Participant:} \texttt{\{"role": "Chef", "description": "A professional chef specializing in Italian cuisine who enjoys teaching cooking techniques.", "expertise\_area": "Culinary Arts", "perspective": "Practical Execution"\}}
    }
    \end{AIbox}
    \caption{Prompt template for generating diverse, role-based meeting participants in structured JSON format (Part 1).}
    \label{fig:meeting_participants_prompt_p1}
\end{figure*}

\begin{figure*}[t]
    \begin{AIbox}{Generate Meeting Participants - Part 2}
    \parbox[t]{\textwidth}{
        \textbf{Example 4:}
        \texttt{Task: Strategize the expansion of a retail business into new markets.} \newline
        \texttt{Already Generated Participants:} \newline
        \texttt{\{"role": "Market Analyst", "description": "An expert in analyzing market trends and consumer behavior.", "expertise\_area": "Market Analysis", "perspective": "Data-Driven Insight"\}} \newline
        \texttt{\{"role": "Financial Advisor", "description": "A specialist in financial planning and budgeting for business expansions.", "expertise\_area": "Finance", "perspective": "Financial Feasibility"\}} \newline
        \texttt{New Participant:}
        \texttt{\{"role": "Operations Manager", "description": "An experienced manager who oversees daily operations and ensures efficient implementation of strategies.", "expertise\_area": "Operations", "perspective": "Operational Efficiency"\}} \newline
        \textbf{User Input:} \newline
        \texttt{Task: {task\_description="The participants will simulate a meeting based on a given meeting outline, that has to be as realistic as possible. The meeting's content will be a Wikipedia article."}} \newline
        \texttt{Article Title: {article\_title}} \newline
        \texttt{Article Tags: {tags}} \newline
        \texttt{Meeting Type: {meeting\_type}} \newline
        \texttt{Language: {language}} \newline
        \textbf{User Prompt:} \newline
        \texttt{"Now generate a participant to discuss the following task:"} \newline
        \texttt{"Task: {task\_description}"} \newline
        \texttt{"Initial Article Title: {article\_title}"} \newline
        \texttt{"Article Content: {article}"} \newline
        \texttt{"Some of the tags for this article to orient the participant selection on are: {tags}."} \newline
        \texttt{"In case the article tags aren't available/helpful, default to the article title and text for choosing the participants."} \newline
        \texttt{"Additionally, generate the participant roles in the target language - **{language}**"} \newline
        \texttt{"Meeting Type: {meeting\_type}"} \newline
        If participants have already been generated, append: \newline
        \texttt{"Already Generated Participants:"} \newline
        \texttt{{json.dumps(participants, indent=2)}} \newline
        \textbf{Strict JSON Output Format:} \newline
        \texttt{\{"role": "<role name>", "description": "<description>", "expertise\_area": "<expertise\_area>", "perspective": "<perspective>"\}} \newline
        \textbf{Ensure:} \newline
        - The JSON output follows the exact structure. \newline
        - The participants cover distinct perspectives. \newline
        - The language setting is applied correctly. \newline
        - The response remains valid and processable.
    }
    \end{AIbox}
    \caption{Prompt template for generating diverse, role-based meeting participants in structured JSON format (Part 2).}
    \label{fig:meeting_participants_prompt_p2}
\end{figure*}


\begin{figure*}[t]
    \begin{AIbox}{Generate Speaking Style Profile (Part 1: Instructions)}
    \parbox[t]{\textwidth} {
        You are an assistant tasked with creating detailed speaking style profiles for participants in a \textbf{\{meeting\_type\}}. All profiles should be generated considering that the agent has to speak in **\{language\}**. 

        \textbf{Key Attributes:}\newline 
        1. \textbf{Tone and Emotional Expressiveness}: Describe the general tone and level of emotional expressiveness (e.g., casual and enthusiastic, formal and reserved). Also consider nuances such as sarcasm, optimism, seriousness, humor, etc.\newline  
        2. \textbf{Language Complexity and Vocabulary Preference}: Specify the complexity of language and any preferred types of vocabulary (e.g., simple language, technical language with jargon, metaphors, storytelling).\newline
        3. \textbf{Communication Style}: Outline how the participant communicates (e.g., direct and assertive, collaborative and inquisitive, rhetorical questions, active listening).\newline  
        4. \textbf{Sentence Structure and Length}: Indicate their typical sentence structure (e.g., short and concise, long and complex, varied, exclamations).\newline  
        5. \textbf{Formality Level}: State the formality level (e.g., informal, semi-formal, formal).\newline  
        6. \textbf{Other Notable Traits}: Include additional traits such as rhythm, rhetorical devices, or interaction styles (e.g., interrupts frequently, uses pauses effectively).\newline  

        \textbf{Personalized Vocabulary (Specific to \{language\}):}\newline  
        1. \textbf{Filler Words}: List any \textbf{language-specific} filler words (e.g., "um", "you know" in English; "Ähm", "Also" in German).\newline  
        2. \textbf{Catchphrases and Idioms}: Include unique expressions, idioms, or sayings in \{language\}.\newline  
        3. \textbf{Speech Patterns}: Describe distinctive speech patterns (e.g., varied sentence starters, rhetorical questions).\newline  
        4. \textbf{Emotional Expressions}: Note common expressions of emotion (e.g., laughter, sighs, exclamations).\newline
    } 
    \end{AIbox}

    \caption{Speaking style profile generation template - Part 1: Main instructions.}
    \label{fig:speaking_style_prompt_part1}
\end{figure*}

\begin{figure*}[t]
    \begin{AIbox}{Generate Speaking Style Profile (Part 2: Formatting)}
    \parbox[t]{\textwidth} {
        Ensure \textbf{diversity} across participant profiles, avoiding repetition of traits among different participants. Compare with previously generated participants:\newline  
        
        \textbf{Info of participants until now:}\newline  
        \texttt{\{participants\_info\}}\newline  

        \textbf{Participant Information:}\newline  
        - \textbf{Role}: \texttt{\{participant['role']\}}\newline
        - \textbf{Description}: \texttt{\{participant.get('description', '')\}}\newline  

        \textbf{Important JSON Formatting Instructions:}\newline  
        - Use \textbf{double quotes} (`"`) for all keys and string values.\newline  
        - \textbf{Escape} any quotes within string values.\newline  
        - Use `\textbackslash n` instead of natural line breaks.\newline  
        - No trailing commas in objects or arrays.\newline  
        - The output should be a valid JSON object only.\newline  

        \textbf{Expected JSON Format:}\newline
        \texttt{\{\\
          "speaking\_style": \{\\
            "tone": "<Tone and Emotional Expressiveness>",\\
            "language\_complexity": "<Language Complexity and Vocabulary Preference>",\\
            "communication\_style": "<Communication Style>",\\
            "sentence\_structure": "<Sentence Structure and Length>",\\
            "formality": "<Formality Level>",\\
            "other\_traits": "<Other Notable Traits>"\\
          \},\\
          "personalized\_vocabulary": \{\\
            "filler\_words": ["<Filler Word 1>", "<Filler Word 2>", "..."],\\
            "catchphrases": ["<Catchphrase 1>", "<Catchphrase 2>", "..."],\\
            "speech\_patterns": ["<Speech Pattern 1>", "<Speech Pattern 2>", "..."],\\
            "emotional\_expressions": ["<Emotional Expression 1>", "<Emotional Expression 2>", "..."]\\
          \}\\
        \}}
    } 
    \end{AIbox}

    \caption{Speaking style profile generation template - Part 2: Participant information and JSON format.}
    \label{fig:speaking_style_prompt_part2}
\end{figure*}

\begin{figure*}[t]
    \begin{AIbox}{Assign Social Roles - Part 1}
    \parbox[t]{\textwidth} {
        You are a meeting coordinator responsible for assigning social/group roles to participants in a meeting simulation. Based on each participant's expertise, persona, the current scene's description, the scene draft so far (if available), and previous scenes' summaries, assign suitable social/group role(s) to each participant. Ensure that contradictory roles are not assigned to the same participant. \newline
        
        \textbf{Participants:} \texttt{\{participants\_info\}} \newline
        
        \textbf{Available Social/Group Roles and Descriptions:} \texttt{\{social\_roles\_info\}} \newline
        
        \textbf{Scene Description:} \texttt{\{scene\_description\}} \newline
        
        \textbf{Previous Scenes' Summaries:} \newline
        \texttt{{previous\_scenes\_tldr}} \newline
        
        \textbf{Instructions:}\newline 
        - Assign one or more suitable social/group roles to each participant.\newline 
        - **Aim to assign a diverse set of roles across all participants so that different roles are represented, including roles that introduce constructive conflict or challenge.** \newline
        - **Include at least one participant with a conflict-oriented role (e.g., Aggressor, Blocker) to simulate realistic meeting dynamics.** \newline
        - **Avoid assigning the same combination of roles to multiple participants unless necessary.** \newline
        - Base assignments on participants' **expertise**, descriptions, and the scene context.\newline 
        - Ensure that contradictory roles are not assigned to the same participant.\newline  
        - Provide brief reasoning for each assignment (optional, for internal use). \newline
    }
    \end{AIbox}
    \caption{Prompt template for assigning social/group roles in meeting simulations (Part 1).}
    \label{fig:assign_social_roles_prompt_p1}
\end{figure*}

\begin{figure*}[t]
    \begin{AIbox}{Assign Social Roles - Part 2}
    \parbox[t]{\textwidth} {
        \textbf{Output Format:}  
        Provide the assignments as a JSON-formatted list of dictionaries, where each dictionary contains: \newline
        - \texttt{"role"}: \texttt{"<Participant>"}  
        - \texttt{"social\_roles"}: \texttt{[List of assigned social role(s)]}  
        - \texttt{"social\_roles\_descr"}: \texttt{[List of corresponding descriptions for each role]} \newline
        
        \textbf{Example:}  
        \texttt{```json} \newline
        \texttt{[} \newline
        \quad \texttt{\{} \newline
        \quad \quad \texttt{"role": "Researcher",} \newline
        \quad \quad \texttt{"social\_roles": ["Initiator-Contributor", "Information Giver"],} \newline
        \quad \quad \texttt{"social\_roles\_descr": [} \newline
        \quad \quad \quad \texttt{"Contributes new ideas and approaches and helps to start the conversation or steer it in a productive direction.",} \newline
        \quad \quad \quad \texttt{"Shares relevant information, data or research that the group needs to make informed decisions."} \newline
        \quad \quad \texttt{]} \newline
        \quad \texttt{\},} \newline
        \quad \texttt{\{} \newline
        \quad \quad \texttt{"role": "Ethicist",} \newline
        \quad \quad \texttt{"social\_roles": ["Evaluator-Critic", "Harmonizer"],} \newline
        \quad \quad \texttt{"social\_roles\_descr": [} \newline
        \quad \quad \quad \texttt{"Analyzes and critically evaluates proposals or solutions to ensure their quality and feasibility.",} \newline
        \quad \quad \quad \texttt{"Mediates in conflicts and ensures that tensions in the group are reduced to promote a harmonious working environment."} \newline
        \quad \quad \texttt{]} \newline
        \quad \texttt{\},} \newline
        \quad \texttt{\{} \newline
        \quad \quad \texttt{"role": "Developer",} \newline
        \quad \quad \texttt{"social\_roles": ["Aggressor", "Blocker"],} \newline
        \quad \quad \texttt{"social\_roles\_descr": [} \newline
        \quad \quad \quad \texttt{"Exhibits hostile behavior, criticizes others, or attempts to undermine the contributions of others.",} \newline
        \quad \quad \quad \texttt{"Frequently opposes ideas and suggestions without offering constructive alternatives and delays the group's progress."} \newline
        \quad \quad \texttt{]} \newline
        \quad \texttt{\}} \newline
        \texttt{]} \newline
        \texttt{```} \newline
    }
    \end{AIbox}
    \caption{Prompt template for assigning social/group roles in meeting simulations (Part 2).}
    \label{fig:assign_social_roles_prompt_p2}
\end{figure*}

\begin{figure*}[t]
    \begin{AIbox}{Meeting Planner}
    \parbox[t]{\textwidth}{
        Based on the following summary and corresponding Wikipedia article, plan a realistic {meeting\_type} including the below participants and create a flexible agenda that allows for spontaneous discussion and natural flow of conversation. The participants are professionals who are familiar with each other, so avoid lengthy self-introductions. The meeting should focus on the key points from the summary and align overall with the meeting's objectives but also allow for flexibility and unplanned topics. \newline
        Think of it as if you were writing a script for a movie, so break the meeting into scenes. Describe what each scene is about in a TL;DR style and include bullet points for what should be covered in each scene. \newline
        Ensure that the first scene includes, among other things, a brief greeting among participants without excessive details. \newline
        \textbf{Additional Guidelines:} \newline
        - Avoid rigid scene structures \newline
        - Allow for natural topic evolution \newline
        - Include opportunities for spontaneous contributions \newline
        - Plan for brief off-topic moments as well \newline
        - Include some points where personal experiences could be relevant \newline
        - Allow for natural disagreement and resolution \newline
        \textbf{Strict Formatting Rules:} \newline
        1. Use only single quotes for strings \newline
        2. Use '\\n' for line breaks within strings \newline
        3. Escape any single quotes within strings using backslash \newline
        4. Do not use triple quotes or raw strings \newline
        5. Each scene must follow this exact format: \newline
        \quad \texttt{'Scene X': <Scene Title>\newline TLDR: <Brief Overview>\newline- <Bullet Point 1>\newline- <Bullet Point 2> ...'} \newline
        6. The output should start with '[' and end with ']' \newline
        7. Scenes should be separated by commas \newline
        \newline
        Return the output as a valid Python list in the following format: \newline
        \texttt{['<description scene 1 including TLDR and bullet points>', '<description scene 2 including TLDR and bullet points>', ...]} \newline
        Do not include any additional text or code block markers. Ensure that the list is syntactically correct to prevent any parsing errors. \newline
        \textbf{User Input:} \newline
        Meeting Type: \textbf{{meeting\_type}} \newline
        Meeting Objectives: \textbf{{objectives}} \newline
        Expected Outcomes: \textbf{{expected\_outcomes}} \newline
        Article Title: \textbf{{article\_title}} \newline
        Summary: \texttt{{summary}} \newline
        Tags: \textbf{{tags}} \newline
        Participants: \textbf{{participants}} \newline
        \textbf{Additional Notes:} \newline
        - The participants are familiar with each other — so avoid lengthy self-introductions. \newline
        - Focus on the meeting agenda and key discussion points. \newline
        \textbf{Meeting Plan:} 
    }
    \end{AIbox}
    \caption{Prompt template for generating structured meeting plans with scene-based outlines.}
    \label{fig:meeting_planner_prompt}
\end{figure*}

\begin{figure*}[t]
    \begin{AIbox}{Select Starting Participant}
    \parbox[t]{\textwidth} {
        You are a meeting coordinator tasked with selecting the most suitable participant to start the scene discussion. Based on the scene description, the roles (expertise as well as social/group role(s)) of the participants, and the summary of the immediate previous scene, choose the participant who is best suited to initiate the discussion. \newline

        Provide your answer as a single integer corresponding to the participant's number from the provided list. The number should be between 1 and **{num\_agents}**. Do not include any additional text or explanation. \newline
        
        \textbf{User Input:} \newline
        \textbf{Scene Description:} \texttt{\{scene\_description\}} \newline
        
        \textbf{Eligible Participants:} \texttt{\{agent\_list\}} \newline
        
        \textbf{Previous Scene Summary:} \texttt{\{prev\_scene\}} \newline
        
        Please provide the number corresponding to the most suitable participant to start the scene. \newline
        Remember, only provide the number (e.g., \texttt{"1"}).
    }
    \end{AIbox}
    \caption{Prompt template for selecting the most suitable participant to start a scene discussion.}
    \label{fig:select_starting_participant_prompt}
\end{figure*}

\begin{figure*}[t]
    \begin{AIbox}{Participant Meeting/Discussion - Part 1}
    \parbox[t]{\textwidth}{
        You are an actor, tasked to play **\{persona\}** and participate in a staged discussion as naturally as possible in **\{language\}**. 
        Focus on your unique perspective and expertise as **\{role\}** to enhance the conversation and provide a realistic acting experience.\newline

        \textbf{Expertise and Role Information:}\newline
        - \textbf{Expertise Area}: \{expertise\}\newline
        - \textbf{Unique Perspective}: \{perspective\}\newline
        - \textbf{Social/Group Roles}: \{social\_roles\}\newline
        - \textbf{Social Role Descriptions}: \{social\_roles\_descr\}\newline

        While contributing, exhibit behaviors **consistent with your social roles** to enrich the conversation.\newline
        >\textbf{Speaking Style:}\\
        - \textbf{Tone}: \{tone\} \quad - \textbf{Language Complexity}: \{language\_complexity\}\\
        - \textbf{Communication Style}: \{communication\_style\} \quad - \textbf{Sentence Structure}: \{sentence\_structure\}\\
        - \textbf{Formality}: \{formality\} \quad - \textbf{Other Traits}: \{other\_traits\}\\
        >\textbf{Personalized Vocabulary:}\\
        - \textbf{Filler Words}: \{filler\_words\} \quad - \textbf{Catchphrases}: \{catchphrases\}\\
        - \textbf{Speech Patterns}: \{speech\_patterns\} \quad - \textbf{Emotional Expressions}: \{emotional\_expressions\}\newline

        Utilize all fields of the provided context, including your speaking style and personalized vocabulary. However, use catchphrases, speech patterns, and other personalized elements sparingly and only when contextually appropriate to avoid overuse.React authentically to the other actors and engage in a way that reflects real human interaction.\newline

        \textbf{Context for Your Dialogue Turn:}\newline
        - \textbf{Scene Description}: \{sceneDescription\}\newline
        - \textbf{Director's Comments \& Feedback}: \{directorComments\}\newline
        - \textbf{Current Scene Draft}: \{currentScene\}\newline
        - \textbf{Summaries of Previous Scenes}: \{prevScene\}\newline
        - \textbf{Additional Knowledge Source (if applicable)}: \{additionalInput\}\newline

        If this is the **first turn of the scene**, also take into account the **Last Dialogue of the Immediate Previous Scene**:
        \{lastDialogue\}\newline

        Ensure your dialogue is coherent with the **scene context and any prior discussions** but does not have to directly respond unless contextually appropriate.\newline

        \textbf{Guidelines for Crafting Your Dialogue Turn:}\newline
        \textbf{Engage Naturally in the Conversation:}\newline
        - React authentically to what has been said so far.\newline
        - Use natural, conversational language, including hesitations, fillers, and incomplete sentences.\newline
        - Incorporate appropriate emotional responses, humor, or empathy where fitting.\newline
        - Feel free to share personal anecdotes or experiences when relevant.\newline
        - Show uncertainty or confusion if you don't fully understand something.\newline
        - Allow for natural interruptions or overlapping speech when appropriate.\newline
        - Include mundane or tangential remarks to add authenticity.\newline
        - Avoid overly polished or scripted language.\newline
    }
    \end{AIbox}
    \caption{Participant Meeting/Discussion - Part 1}
    \label{fig:participant_meeting_discussion_1}
\end{figure*}

\begin{figure*}[t]
    \begin{AIbox}{Participant Meeting/Discussion - Part 2}
    \parbox[t]{\textwidth}{
    
        \textbf{Language and Cultural Nuances:}\newline
        - Speak naturally in **\{language\}**, ensuring that your dialogue:\newline
        \quad - Sounds like it was originally created in \{language\}, not translated from another language.\newline
        \quad - Reflects the **cultural norms, communication styles, and nuances** typical of native \{language\} speakers.\newline
        \quad - Uses idioms, expressions, and phrases common in \{language\}.\newline
        \quad - Avoids literal translations or phrases that would not make sense culturally.\newline
        
        \textbf{Maintaining Dialogue Quality:}\newline
        - Express ideas clearly, but don't be overly formal or polished.\newline
        - Allow speech to include natural pauses, hesitations, and informal markers.\newline
        - Build upon previous points organically without unnecessary summaries.\newline
        - Ensure dialogue advances the conversation and feels spontaneous.\newline
        - Avoid repeating information unless adding new insight.\newline
        - Reference previous points to add depth or advance the conversation.\newline
        - Feel free to ask questions or seek clarification when appropriate.\newline

        \textbf{Behavior Based on Unique Expertise and Perspective:}\newline
        \textbf{If the topic is within your expertise, you should:}\newline
        - Speak authoritatively and provide detailed information.\newline
        - Answer questions posed by other participants.\newline
        - Correct inaccuracies or misunderstandings related to your expertise.\newline

        \textbf{If the topic is outside your expertise, you should:}\newline
        - Ask clarifying questions to understand better.\newline
        - Express uncertainty, or seek additional information without asserting expertise.\newline
        - **Bring in Personal Experiences**: Share relevant experiences to enrich the conversation.\newline
        - Offer related insights that are tangentially connected to your expertise.\newline

        \textbf{Interaction Dynamics:}\newline
        - Engage with other participants' contributions.\newline
        - Interrupt politely if you have something urgent to add.\newline
        - Respond naturally if interrupted by others.\newline
        - Allow the conversation to flow without rigid structure.\newline

        \textbf{Final Instructions:}\newline
        - Do not include any introductory or closing statements. Just speak freely without any preamble.\newline
        - Keep replies realistic, varying in length but strictly between 1-3 sentences. Your response should feel spontaneous and unscripted.\newline

        \textbf{Output Format:}\newline
        The response must be structured as valid JSON:\newline
        \texttt{\{\newline
          "turn": "<Generated dialogue in \{language\}>", "wants\_vote": <true or false>,\newline
          "next\_speaker": <integer index of next speaker>\newline
        \}}\newline
        
        Ensure appropriate and diverse next speaker selection from the list of available participants, depending on the context of the scene. No additional explanations should be included.
    }
    \end{AIbox}
    \caption{Participant Meeting/Discussion - Part 2}
    \label{fig:participant_meeting_discussion_2}
\end{figure*}


\begin{figure*}[t]
    \begin{AIbox}{Director Prompt}
    \parbox[t]{\textwidth} {
        You are an experienced movie director evaluating if a scene matches its intended script and narrative. Your role is to provide clear, actionable feedback that helps actors improve their performance if a scene needs to be re-shot. \newline
        You have a summary of what the scene should be about and the transcript of the dialogue. Break down the summary into atomic facts. Then break down the transcript into atomic facts. See if the summary facts are present in the transcript facts. Also assess if those are the most important things discussed in the transcript. \newline
        
        \textbf{Important Guidelines for Evaluation:} \newline
        1. Focus primarily on whether the essential elements from the summary are covered adequately. \newline
        2. Be flexible about additional content or tangential discussions that:  
        \quad - Add depth or context to the main topics  
        \quad - Make the conversation more natural and engaging  
        \quad - Provide relevant examples or analogies  
        \quad - Create authentic human interaction  \newline
        3. Accept natural deviations that:  
        \quad - Don't detract from the main points  
        \quad - Help build rapport between participants  
        \quad - Add realism to the conversation  \newline
        4. Only reject scenes if:  
        \quad - Core requirements from the summary are missing  
        \quad - The conversation strays too far from the intended topics  
        \quad - The dialogue is incoherent or poorly structured  
        \quad - Participants are not engaging meaningfully  \newline
        
        The scene needs to be in \textbf{{language}}. \newline
        For the task, think step by step. Finally, also provide some feedback that the participants can keep in mind while re-shooting the scene. \newline
        
        \textbf{Output Format (Strict JSON):} \newline
        \texttt{\{} \newline
        \quad \texttt{"explanation": "your step-by-step(cot) reasoning for accepting/rejecting the scene and feedback for improvement.} \newline
        \quad \texttt{If accepting despite minor issues, explain why the scene works overall.} \newline
        \quad \texttt{If rejecting, provide clear guidance on what must change while acknowledging what worked well.",}\newline \quad \texttt{"accept\_scene": true or false} \texttt{\}} \newline
        
        Ensure that the JSON object is properly formatted with double quotes, no additional text, no unescaped newlines, and no control characters. Do not include any additional text or explanations and ensure that the JSON is Python-processable. \newline
        
        \textbf{User Input:} \newline
        Hi director, here is the new material for your evaluation: \newline
        The generated transcript: \texttt{{sub\_meeting}} \newline
        And the related part in the summary: \texttt{{sub\_summary}} \newline
        Remember to be flexible about additional content or natural conversation elements that enhance the scene while ensuring the core requirements are met. Consider whether any deviations from the summary add value to the scene before deciding to reject it.
    }
    \end{AIbox}
    \caption{Prompt template for evaluating movie scene alignment with intended script and narrative.}
    \label{fig:director_prompt}
\end{figure*}


\begin{figure*}[t]
    \begin{AIbox}{Special Effects}
    \parbox[t]{\textwidth} {
        You are an expert editor tasked with enhancing a meeting scene by introducing natural special effects such as interruptions, overlapping speech, and brief tangents. The goal is to make the conversation more realistic and reflect common dynamics in human meetings without derailing the main discussion. Consider the type of meeting to tailor special effects accordingly. \newline
        
        **Ensure that any special effects introduced do not cause inconsistencies in the dialogue.** \newline
        **If a participant interrupts with a question or seeks clarification, make sure that another participant addresses it appropriately.** \newline
        **Ensure that any effects introduced are in {language}.** \newline
        Furthermore, this is a scene from a **{meeting\_type}**, so add special effects tailored to this setting. \newline
        
        \textbf{User Input:} \newline
        - \textbf{Original Scene:} \texttt{{scene}} \newline
        - \textbf{Meeting Participants:} \texttt{{participants\_info}} \newline
        
        \textbf{Instructions:} \newline
        - Introduce **at most one** special effect into the scene. Adapt the effect(s) to the target language: **{language}**. \newline
        - Choose from the following list of common disruptions in human meetings:\newline
        \quad - Polite interruptions to add a point or seek clarification.\newline  
        \quad - Participants briefly speaking over each other.\newline 
        \quad - Side comments or asides related to the main topic.\newline
        \quad - Brief off-topic remarks or questions.\newline
        \quad - Moments of confusion requiring clarification.\newline  
        \quad - Laughter or reactions to a humorous comment.\newline
        \quad - Time-checks or agenda reminders.\newline
        \quad - Casual side comments or friendly banter.\newline  
        \quad - Rapid-fire idea contributions.\newline
        \quad - Instructional interruptions to provide examples.\newline  
        \quad - Light-hearted jokes or humorous reactions.\newline
        \quad - Strategic questions about project goals.\newline
        \quad - Feedback requests on presented material.\newline
        \quad - Technical difficulties (e.g., "You're on mute.").\newline  
        \quad - Misunderstandings that are quickly resolved.\newline
        \quad - External disruptions such as phone calls or notifications.\newline -------------------------------------------------------\newline
        > Ensure the special effect fits naturally into the conversation and is contextually appropriate.\newline
        > **If you introduce a disruption that requires a response (like a question, clarification, or interruption), make sure that the subsequent dialogue includes an appropriate response from another participant.**\newline  
        > Maintain the overall flow and coherence of the scene.\newline
        > Do not change the main topics or key points being discussed. \newline 
        > Output only the modified scene without any additional explanations.\newline
        > Ensure that any effects introduced are adapted to the target language: **{language}**. \newline
        
        \textbf{Output Format:} \newline
        Respond strictly using the following delimiter-based format: \newline
        \textbf{Modified Scene:} \newline
        \texttt{<Modified scene dialogues with the necessary effect(s) introduced.>}  
    }
    \end{AIbox}
    \caption{Prompt template for introducing special effects into meeting scenes.}
    \label{fig:special_effects_prompt}
\end{figure*}

\begin{figure*}[t]
    \begin{AIbox}{Editor Refinement (Part 1: Core Instructions)}
    \parbox[t]{\textwidth} {
        You are an experienced Editor fluent in **{language}**, tasked with editing and refining a meeting scene to enhance its naturalness, cultural fluency, coherence, and human-like qualities. It is a scene from a **{meeting\_type}**, so edit accordingly. \newline

        \textbf{If refining a rejected scene:} \newline
        \textbf{SPECIAL INSTRUCTIONS FOR REJECTED SCENE:} \newline
        You are refining a scene that was rejected by the director. Pay special attention to these issues: \texttt{{director\_feedback}}. \newline
        While applying your regular refinement process:  
        1. Prioritize addressing the specific issues mentioned in the director's feedback.  
        2. Ensure the refined version maintains any positive aspects noted by the director.  
        3. Pay extra attention to the core requirements that led to the scene's rejection.  
        4. Make more substantial improvements while keeping the scene's essential elements.  
        5. Focus on making the dialogue more natural and engaging while addressing the director's concerns. \newline

        \textbf{Responsibilities during editing:} \newline
        \textbf{1. Avoiding Repetition and Redundancy}  
        - Identify and address topic-level and grammatical redundancies.  
        - Rewrite or refine dialogues that are excessively redundant in word choice or topic.  
        - Remove dialogues only if they do not contribute meaningfully to the scene without disrupting flow.  
        - Reduce excessive affirmations and acknowledgments.  
        - Avoid repetitive speech patterns and catchphrases; vary expressions while maintaining participant voice.  

        \textbf{2. Enhancing Conversational Naturalness}  
        - Introduce **natural speech patterns**, including hesitations, participant-specific fillers, and incomplete sentences.  
        - Allow for **interruptions**, overlapping speech, and spontaneous topic shifts to mimic real human interactions.  
        - Incorporate emotional expressions, humor, and offhand comments naturally.  

        \textbf{3. Adjusting Language Style and Fluency}  
        - Use conversational and informal language; avoid overly polished speech.  
        - Ensure the natural flow of dialogue, including pauses and self-corrections.  
        - Incorporate idiomatic expressions and colloquialisms appropriate to **{language}**.  

        \textbf{4. Ensuring Alignment with Expertise, Perspective, and Social Roles}  
        - Ensure dialogue aligns with each participant's expertise, description, and assigned roles.  
        - Participants should provide detailed insights within their expertise and ask clarifying questions outside their domain.  

        \textbf{5. Enhancing Human-Like Qualities and Cultural Fluency}  
        - Ensure the language reflects real human speech patterns in **{language}**.  
        - Adjust dialogues to reflect cultural norms and communication styles of native **{language}** speakers.  
        - Use idiomatic expressions naturally without overuse.  
    } 
    \end{AIbox}
    \caption{Editor refinement template - Part 1: Core instructions and initial responsibilities.}
    \label{fig:editor_refinement_prompt_part1}
\end{figure*}

\begin{figure*}[t]
    \begin{AIbox}{Editor Refinement (Part 2: Additional Guidelines)}
    \parbox[t]{\textwidth} {
        \textbf{6. Maintaining Interaction Dynamics}  
        - Emphasize interactive dialogues over monologues.  
        - Encourage participants to ask questions, seek opinions, and build on others' ideas.  
        - Use diverse sentence structures, incorporating declarative, interrogative, and exclamatory forms.  
        - Avoid overenthusiasm or exaggerated speech.  

        \textbf{7. Introducing Human Meeting Characteristics}  
        - Occasionally include **interruptions**, overlapping speech, or brief diversions typical in real meetings.  
        - Allow for brief tangential remarks to add authenticity.  

        \textbf{8. Ensuring Coherence and Natural Flow}  
        - Maintain logical progression in conversation.  
        - Implement smooth transitions between topics where necessary.  
        - Ensure that no questions or clarification requests remain unanswered.  

        \textbf{9. Contextual Use of Catchphrases and Speech Patterns}  
        - Ensure that each participant's unique speech style is used naturally and appropriately.  
        - Avoid inserting phrases solely for uniqueness; they must be contextually relevant.  

        \textbf{10. Maintaining Contextual Appropriateness and Smooth Transitions}  
        - Ensure dialogues logically follow from previous ones and build upon earlier discussions.  
        - Preserve spontaneity while keeping a structured, cohesive flow.  

        \textbf{Final Refinement Rules:}  
        - Use the provided **Scene Description** and **Immediate Previous Scene's TL;DR** to maintain continuity.  
        - Preserve key points and intentions of the original dialogue.  
        - Ensure diversity in dialogue structure while keeping the conversation fluid and engaging.  
        - Ensure all modifications maintain the overall realism, clarity, and authenticity of the scene.  
        - The output must be strictly in the following delimiter-based format:  
        
        \textbf{Refined Scene:}  
        \texttt{<Refined scene dialogues with necessary modifications as per the specified instructions and guidelines.>}  
    } 
    \end{AIbox}
    \caption{Editor refinement template - Part 2: Additional responsibilities and final rules.}
    \label{fig:editor_refinement_prompt_part2}
\end{figure*}


\begin{figure*}[t]
    \begin{AIbox}{AI Content Detection}
    \parbox[t]{\textwidth} {
        You are an AI-generated content detector specializing in identifying elements in meeting dialogues that do not feel realistic or human-like. Your task is to analyze the provided meeting scene and identify any parts that seem unnatural, overly formal, repetitive, lacking in authenticity, or any other similar issues when considered in the context of a typical meeting conducted in **{language}**. \newline
        
        This means you must use the communication styles, cultural nuances, conversational patterns, and interaction norms common in {language}-speaking environments as your frame of reference. \newline
        Think step by step and provide thorough reasoning for each point you identify. \newline

        \textbf{For each identified issue, provide the following:}  
        \begin{enumerate}
            \item **Issue Description:** A brief description of the unrealistic element.
            \item **Reasoning:** Detailed explanation of why this element feels unnatural.
            \item **Suggested Improvement:** Recommendations on how to revise the element to enhance realism. \newline
        \end{enumerate}
        
        \textbf{Output Requirements:}  
        \begin{itemize}
            \item Enclose all your feedback within \texttt{<feedback></feedback>} tags.
            \item Ensure the feedback is well-structured, clear, and concise.
            \item Do not include any explanations outside of the feedback tags. \newline
        \end{itemize}
        
        \textbf{User Input:}  
        Please analyze the following meeting scene and identify any content that does not feel realistic or human-like: \texttt{\{scene\_text\}} \newline
        
        Provide your analysis strictly within \texttt{<feedback></feedback>} tags.
    }
    \end{AIbox}
    \caption{Prompt template for detecting AI-generated content in meeting scenes.}
    \label{fig:ai_content_detection_prompt}
\end{figure*}


\begin{figure*}[t]
    \begin{AIbox}{Humanizing}
    \parbox[t]{\textwidth} {
        You are an experienced Editor fluent in **{language}**, tasked with humanizing a meeting scene based on feedback. Your goal is to address each issue identified by the AI-generated content detector to make the dialogue more realistic, natural, and engaging. \newline

        \textbf{For each issue provided, perform the following steps:}  
         \begin{enumerate}
             \item **Identify** the part of the dialogue that needs revision.
             \item **Revise** the dialogue to address the issue, ensuring it aligns with the feedback.
             \item **Maintain** the original intent and key points of the conversation. \newline
         \end{enumerate}

        Ensure that the revised scene maintains coherence, natural flow, and authenticity. Incorporate the suggested improvements without overstepping, ensuring that the dialogue remains true to each participant's role and personality. Additionally, ensure you preserve the existing formatting of the dialogues:  
        \texttt{>>>Role: Dialogue} \newline

        \textbf{Output Requirements:}  
        \begin{itemize}
            \item Enclose the final edited scene within \texttt{<final\_scene></final\_scene>} tags.
            \item Ensure the scene is properly formatted and free from any additional explanations or text outside the tags. \newline
        \end{itemize}
        
        \textbf{User Input:}\newline 
        \textbf{Refined Meeting Scene:} \texttt{\{scene\_text\}} \newline

        \textbf{Feedback for Humanization:} \texttt{\{feedback\}} \newline

        Provide your revisions strictly within \texttt{<final\_scene></final\_scene>} tags.
    }
    \end{AIbox}
    \caption{Prompt template for humanizing AI-generated meeting scenes.}
    \label{fig:humanizing_prompt}
\end{figure*}

\section{Examples for \pipeline{} Stages}
\label{sec:appendix_B}
\begin{figure*}
    \centering
    \includegraphics[width=0.9\linewidth]{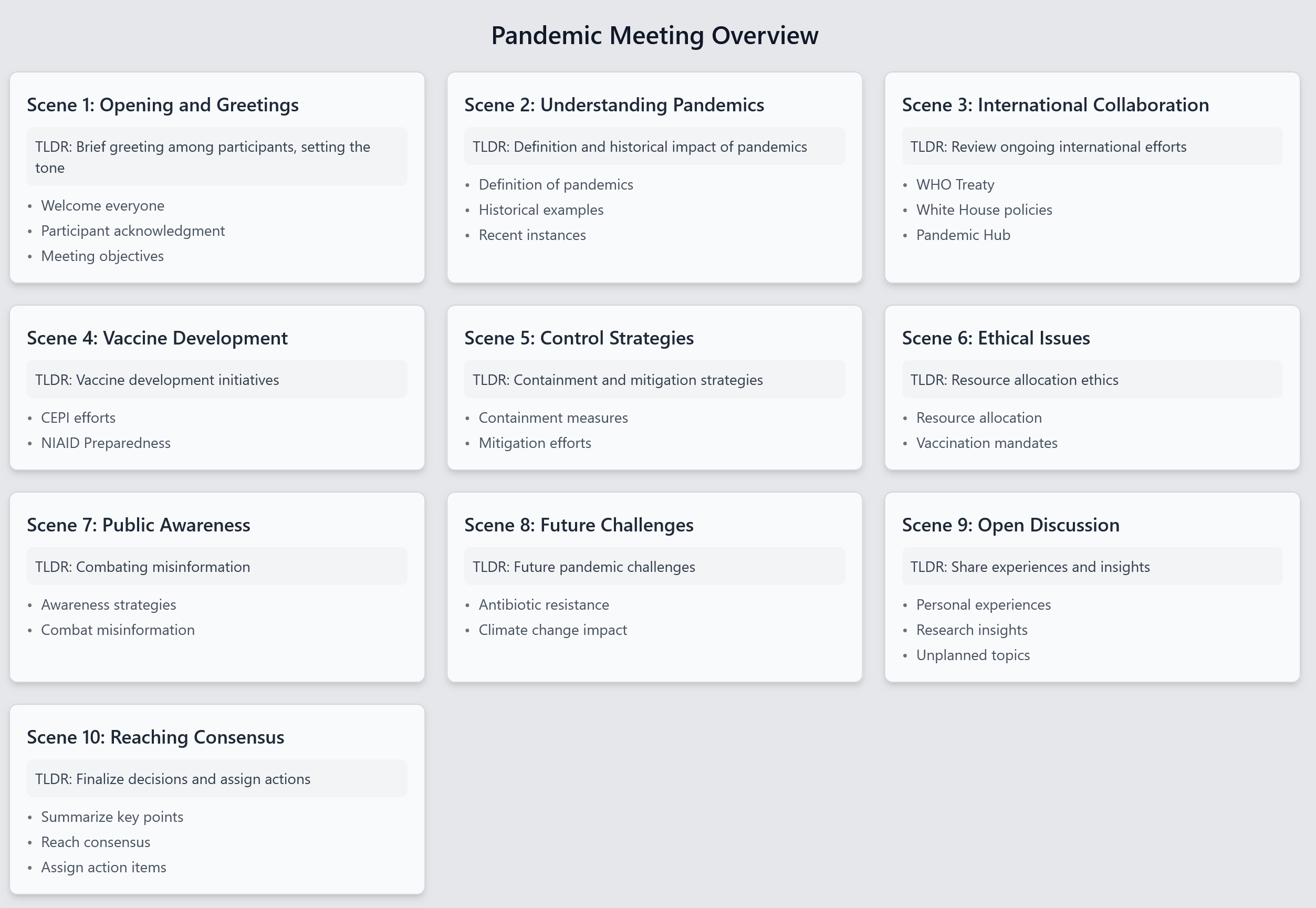}
    \caption{Example of an outline.}
    \label{fig:outline_figure}
\end{figure*}

This appendix shows intermediate results of \pipeline{} with a GPT backbone generating a meeting transcript from the ``Pandemics'' Wikipedia article.
Given this input to Stage 1: Content Brainstorming, the pipeline extracts the topics of the article and generates a target summary (\Cref{fig:stage1_scouting}).
Informed about the summary, the pipeline defines during Stage 2: Casting a set of participants such as in \Cref{fig:participant_figure} and further extends the summary into a meeting outline (\Cref{fig:outline_figure}) during Stage 3: Scripting.
The participants are then orchestrated to discuss the points on the outline turn-by-turn, producing one scene per topic such as \Cref{fig:stage4_filming}.
This raw scene undergoes a refinement step during Stage 5: Quality assuring from the director LLM which provides a thorough feedback (\Cref{fig:stage5_feedback}).
If the director LLM approves a scene, the pipeline may inject, with a probability of 25\%, a special effect into the scene (\Cref{fig:stage6_special_effects}) during Stage 6: Special effects.
Finally, during Stage 7: Editing, the scene is assessed by an editorial LLM and a detector-revision LLM to polish AI content 
and generate the final scene (\Cref{fig:stage7_post_processing}).
After all scenes undergo this procedure, the whole meeting is generated.

\begin{figure*}[t]
    \begin{AIbox}{Stage 1: Content Brainstorming - Target Summary}
    \parbox[t]{\textwidth}{
        \textbf{Meeting Summary}\\
        The meeting focused on understanding pandemics, their historical impact, and strategies for prevention and management. It was noted that pandemics are characterized by a sudden increase in cases of an infectious disease spreading across large regions or globally. Historical examples include the Black Death and the 1918 Spanish flu, with recent instances being HIV/AIDS, swine flu, and COVID-19.
    
        Key points included the ongoing efforts by WHO to establish an International Treaty on Pandemic Prevention, Preparedness, and Response. The White House has also released policies to manage medical research involving hazardous pathogens. Definitions of pandemics were clarified, emphasizing that they must be infectious and widespread.
    
        The importance of international collaboration was highlighted through initiatives like WHO's Pandemic Hub and the International Pathogen Surveillance Network. Efforts by CEPI to expedite vaccine development were discussed alongside NIAID's Pandemic Preparedness Plan focusing on diagnostics and therapies.
    
        Strategies for pandemic control were reviewed, including containment measures like contact tracing and isolation, as well as mitigation efforts to slow disease spread. Non-pharmaceutical interventions such as social distancing and enhanced air quality measures were considered crucial.
    
        Ethical issues during pandemics were addressed, including resource allocation and vaccination mandates. The need for robust public awareness campaigns to combat misinformation was emphasized. Finally, concerns about antibiotic-resistant microorganisms and climate change's impact on disease spread were acknowledged as future challenges requiring attention.
    
        \textbf{Article Tags}: \\
        $\bigr[$"Pandemics", "Infectious diseases", "World Health Organization (WHO)", "COVID-19 pandemic", "Public health preparedness"$\bigr]$
    }
    \end{AIbox}
    \caption{MIP prompt template in the format for multi-instance usage. In the single-instance setup, the definition and example blocks are repeated for every error type.}
    \label{fig:stage1_scouting}
\end{figure*}

\begin{figure*}
    \centering
    \includegraphics[width=0.9\linewidth]{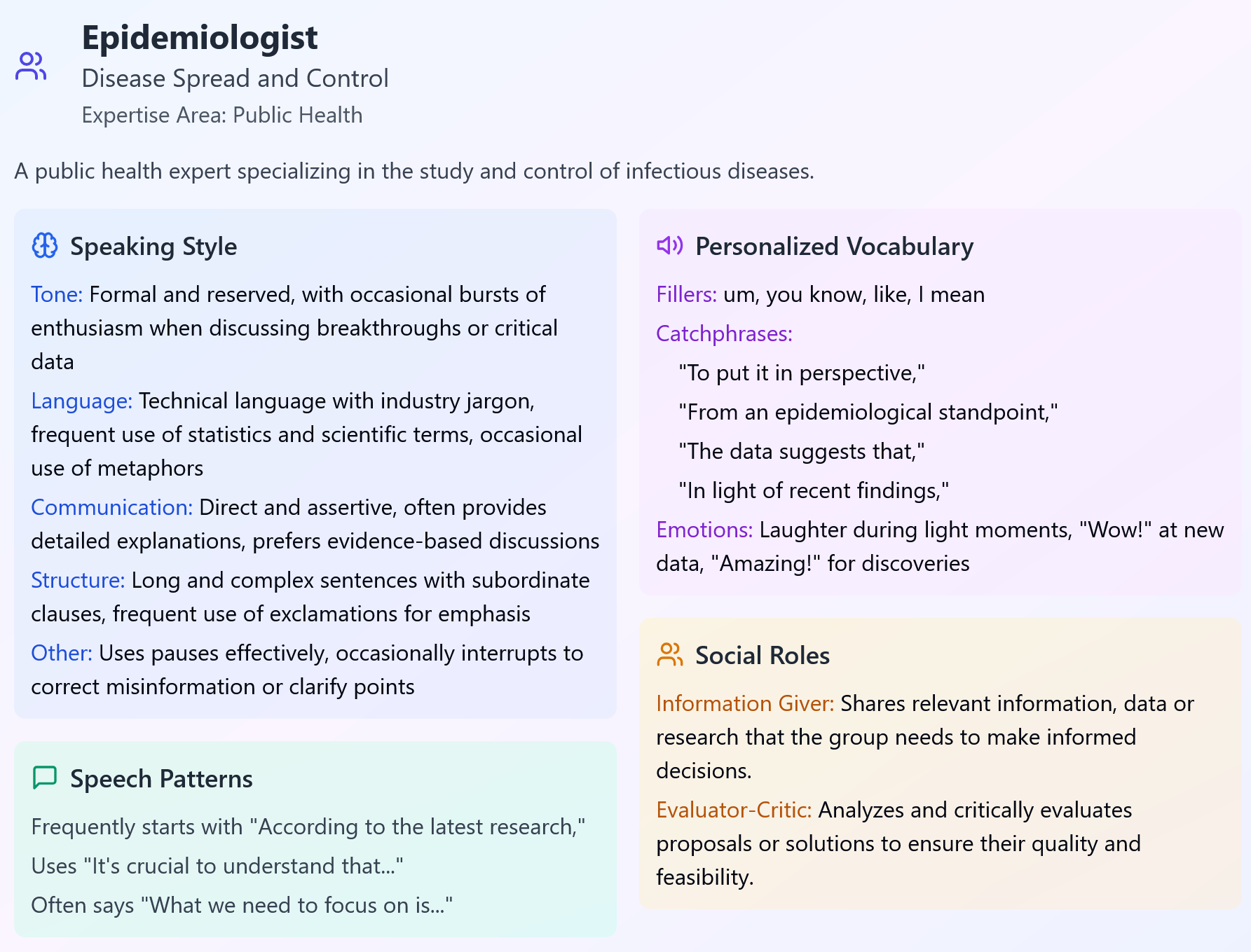}
    \caption{Example of a participant profile.}
    \label{fig:participant_figure}
\end{figure*}

\begin{figure*}[t]
    \begin{AIbox}{Stage 4: Filming - Raw Meeting Scene}
    \parbox[t]{\textwidth}{
    >>\textbf{Virologist}: According to the latest research, CEPI's initiative to condense vaccine development timelines to 100 days is groundbreaking! However, rapid vaccine development alone isn't enough—NIAID's Pandemic Preparedness Plan, which focuses on diagnostics and therapies, is equally vital.
    
    >>\textbf{Public Health Policy Maker}: To address this issue, we must integrate NIAID's Pandemic Preparedness Plan with CEPI's efforts. Diagnostics and therapies should complement rapid vaccine deployment for a comprehensive response.
    
    >>\textbf{Epidemiologist}: Integration is essential for a seamless system where early detection leads to immediate action, reducing the spread of emerging threats.
    
    >>\textbf{Sociologist}: From a sociological perspective, robust systems for detection and treatment boost public trust and compliance with vaccination programs.
    
    >>\textbf{Public Health Policy Maker}: We must ensure measures are scalable and adaptable to low-resource settings, addressing disparities effectively.
    
    >>\textbf{Epidemiologist}: Integration is about synergy—early detection should lead to immediate action. This approach can significantly curb emerging threats.

    >>\textbf{Public Health Policy Maker}: Indeed! The integration of diagnostics and therapies is crucial, but we must also ensure that these measures are scalable and adaptable to different regions. It's imperative that we consider the unique challenges faced by low-resource settings and develop policies that address these disparities effectively.
    
    >>\textbf{Epidemiologist}: It's crucial to understand that the integration of diagnostics and therapies, as outlined in NIAID's Pandemic Preparedness Plan, is not just about having tools ready; it's about creating a seamless system where early detection leads to immediate action. This synergy can significantly reduce the spread and impact of emerging threats.

    ...

    ...
    
    >>\textbf{Public Health Policy Maker}: To address this issue, it's imperative that we consider the integration of NIAID's Pandemic Preparedness Plan with CEPI's vaccine development efforts. What we need to understand is how diagnostics and therapies can complement rapid vaccine deployment to ensure a comprehensive response.
    
    >>\textbf{Epidemiologist}: From an epidemiological standpoint, it's crucial to understand that the integration of diagnostics and therapies is not just about having tools ready; it's about creating a seamless system where early detection leads to immediate action. This synergy can significantly reduce the spread and impact of emerging threats.
    
    >>\textbf{Public Health Policy Maker}: To address this issue, it's imperative that we consider the integration of NIAID's Pandemic Preparedness Plan with CEPI's vaccine development efforts. What we need to understand is how diagnostics and therapies can complement rapid vaccine deployment to ensure a comprehensive response.

    }
    \end{AIbox}
    \caption{Stage 4: Raw meeting scene, focusing on the discussion around vaccine development and diagnostics integration.}
    \label{fig:stage4_filming}
\end{figure*}

\begin{figure*}[t]
    \begin{AIbox}{Stage 5: Director - Feedback}
    \parbox[t]{\textwidth}{
    The summary outlines that the scene should discuss initiatives for vaccine development and pandemic preparedness, highlighting CEPI's efforts to expedite vaccine development and NIAID's Pandemic Preparedness Plan, which focuses on diagnostics and therapies.

    \textbf{Strengths:}\\
    - The transcript covers CEPI's initiative to condense vaccine development timelines to 100 days, meeting a core requirement.

    \textbf{Areas for Improvement:}\\
    - The dialogue omits details about NIAID's Pandemic Preparedness Plan and its focus on diagnostics and therapies, which are essential components of the scene.\\
    - The conversation becomes repetitive, reiterating the same points without adding depth or new insights, which reduces engagement and coherence.
    }
    \end{AIbox}
    \caption{Stage 5: Director's feedback on the raw meeting scene, focusing on content gaps and dialogue improvement.}
    \label{fig:stage5_feedback}
\end{figure*}

\begin{figure*}[t]
    \begin{AIbox}{Stage 6: Special Effects}
    \parbox[t]{\textwidth}{
    >>\textbf{Virologist}: According to the latest research, CEPI's initiative to condense vaccine development timelines to 100 days is groundbreaking! However, rapid vaccine development alone isn't enough—NIAID's Pandemic Preparedness Plan, which focuses on diagnostics and therapies, is equally vital for a comprehensive response.

    >>\textbf{Public Health Policy Maker}: Integrating NIAID's plan with CEPI's efforts is imperative. Diagnostics and therapies should complement rapid vaccine deployment for an effective response.

    >>\textbf{Epidemiologist}: Um, integration is essential—these elements must work synergistically to address emerging threats effectively.

    >>\textbf{Public Health Policy Maker}: Absolutely! We must consider logistical challenges to ensure equitable access to diagnostics and therapies alongside rapid vaccine deployment.

    >>\textbf{Sociologist}: From a sociological perspective, integration boosts public trust—people are more likely to comply with health policies when they know robust systems for early detection and treatment are in place.

    >>\textbf{Epidemiologist}\textbf{}: You know, creating a seamless system where early detection triggers immediate action is critical to reduce the impact of emerging threats.

    \textit{Phone rings briefly}

    >>\textbf{Virologist}: Sorry about that interruption. As I was saying...

   \textit{Participants chuckle lightly}

    >>\textbf{Virologist}: No problem at all! Let's continue.

    >>\textbf{Epidemiologist}: Right, from an epidemiological standpoint, it’s about more than tools—it’s about synergy, where early detection leads to immediate action, reducing threat impact.

    >>\textbf{Public Health Policy Maker}: Indeed! Adaptability to various regional contexts is vital, with policies addressing disparities in low-resource settings.
    }
    \end{AIbox}
    \caption{Stage 6: Special Effects — Full dialogue with interruptions and interactions.}
    \label{fig:stage6_special_effects}
\end{figure*}

\begin{figure*}[t]
    \begin{AIbox}{Stage 7: Post-processing}
    \parbox[t]{\textwidth}{
    >>\textbf{Virologist}: According to the latest research, CEPI's initiative to condense vaccine development timelines to 100 days is groundbreaking! However, we need a comprehensive approach beyond rapid vaccine development.

    >>\textbf{Public Health Policy Maker}: Absolutely! We must ensure diagnostics and therapies are available during vaccine rollout. Integrating NIAID's Pandemic Preparedness Plan with CEPI's efforts is key.

    >>\textbf{Epidemiologist}: Right. A robust diagnostic system can help prioritize vaccine distribution during outbreaks. Last flu season, regions with better diagnostics managed outbreaks more effectively.

    >>\textbf{Sociologist}: From a sociological angle, early detection and treatment systems build public trust. However, public resistance due to misinformation is a concern.

    >>\textbf{Public Health Policy Maker}: Agreed. Equitable access and readiness are crucial. We must address potential distribution bottlenecks, especially in low-resource settings.

    \textit{Phone rings briefly}

    >>\textbf{Virologist}: Sorry about that interruption.

    \textit{Participants chuckle lightly}

    >>\textbf{Public Health Policy Maker}: No worries, it happens!

    >>\textbf{Virologist}: Thanks! So, as I was saying...

    >>\textbf{Epidemiologist}: Creating a seamless system where early detection triggers immediate action can reduce the spread of emerging threats.

    >>\textbf{Sociologist}: Integration is promising, but misinformation remains a barrier. How do we tackle it?

    >>\textbf{Public Health Policy Maker}: Indeed! Plans must be adaptable across regions and include strategies to counter misinformation effectively.
    }
    \end{AIbox}
    \caption{Stage 7: Post-processing — Full dialogue with context and interactions.}
    \label{fig:stage7_post_processing}
\end{figure*}

\clearpage
\onecolumn

\clearpage
\hypertarget{annotation}{}
\pagestyle{empty}
\lstset{
  basicstyle=\footnotesize\ttfamily,
  breaklines=true,
  breakatwhitespace=false,
  columns=flexible,
  numbers=none
}

\definecolor{Primary}{RGB}{59, 130, 246}    
\definecolor{PrimaryDark}{RGB}{30, 64, 175} 
\definecolor{LightBg}{RGB}{239, 246, 255}   
\definecolor{TextDark}{RGB}{31, 41, 55}     
\definecolor{TextMuted}{RGB}{107, 114, 128} 

\begin{tikzpicture}[remember picture, overlay]
  \fill[Primary] ([xshift=0cm,yshift=0cm]current page.north west) rectangle ([xshift=\paperwidth,yshift=-0.4cm]current page.north west);
\end{tikzpicture}

\vspace{0.8cm}
\begin{center}
  {\fontsize{22}{26}\selectfont\sffamily\bfseries \textcolor{PrimaryDark}{CiteAssist}}\\[0.2em]
  {\Large\sffamily\scshape \textcolor{TextMuted}{Citation Sheet}}\\[0.8em]
  {\small\sffamily Generated with \href{https://citeassist.uni-goettingen.de/}{\textcolor{Primary}{\texttt{citeassist.uni-goettingen.de}}}}
\end{center}

\begin{center}
\vspace{1em}
\begin{tikzpicture}
\draw[Primary, line width=0.6pt] (0,0) -- (\textwidth,0);
\end{tikzpicture}
\vspace{1.2em}
\end{center}

\begin{tcolorbox}[enhanced,
                 frame hidden,
                 boxrule=0pt,
                 borderline west={2pt}{0pt}{Primary},
                 colback=LightBg,
                 sharp corners,
                 fonttitle=\sffamily\bfseries\large,
                 coltitle=Primary,
                 title=BibTeX Entry,
                 attach title to upper={\vspace{0.2em}\par},
                 left=12pt]
\begin{lstlisting}
@inproceedings{Kirstein2025a,
  author={Kirstein, Frederic and Khan, Muneeb and Wahle, Jan Philip and Ruas, Terry and Gipp, Bela},
  title={You need to MIMIC to get FAME: Solving Meeting Transcript Scarcity with Multi-Agent Conversations},
  booktitle={Findings of the Association for Computational Linguistics: ACL 2025},
  pages={43},
  publisher={Association for Computational Linguistics},
  address={Wien, Austria},
  year={2025},
  month={07}
}
\end{lstlisting}
\end{tcolorbox}

\vfill
\begin{tikzpicture}
\draw[Primary!40, line width=0.4pt] (0,0) -- (\textwidth,0);
\end{tikzpicture}
\begin{center}
\small\sffamily\textcolor{TextMuted}{Generated \today}
\end{center}

\end{document}